\documentclass[accepted]{uai2022} 

\usepackage[american]{babel}

\usepackage{natbib} 
\usepackage{mathtools} 
\usepackage{booktabs} 
\usepackage{tikz} 

\usepackage{algorithm}
\usepackage[compatible]{algpseudocode} 

\usepackage{amsmath}

\DeclareMathOperator*{\argmin}{arg\,min}
\usepackage{bm}
\usepackage{amssymb}

\usepackage{mathtools}
\usepackage{amsthm}
\usepackage{color}
\usepackage{microtype}
\usepackage{graphicx}
\usepackage{subfigure}
\usepackage{arydshln}

\usepackage{multirow}

\makeatletter
\def\adl@drawiv#1#2#3{%
        \hskip.5\tabcolsep
        \xleaders#3{#2.5\@tempdimb #1{1}#2.5\@tempdimb}%
                #2\z@ plus1fil minus1fil\relax
        \hskip.5\tabcolsep}
\newcommand{\cdashlinelr}[1]{%
  \noalign{\vskip\aboverulesep
           \global\let\@dashdrawstore\adl@draw
           \global\let\adl@draw\adl@drawiv}
  \cdashline{#1}
  \noalign{\global\let\adl@draw\@dashdrawstore
           \vskip\belowrulesep}}
\makeatother

\title{Superposing Many Tickets into One: \\ A Performance Booster for Sparse Neural Network Training}

\author[1]{Lu Yin}
\author[1]{Vlado Menkovski}
\author[1]{Meng Fang}
\author[1]{Tianjin Huang}
\author[1]{Yulong Pei}
\author[1]{Mykola Pechenizkiy} 
\author[2,1]{\\Decebal Constantin Mocanu}
\author[1]{Shiwei Liu\thanks{Corresponding author: Shiwei Liu, s.liu3@tue.nl}} 
\affil[1]{%
    Eindhoven University of Technology \\
    Eindhoven, the Netherlands
}
\affil[2]{%
    University of Twente\\
    Enschede, the Netherlands
}

\begin{document}
\maketitle

\begin{abstract}
\vspace{-1em}
Recent works on sparse neural network training (sparse training) have shown that a compelling trade-off between performance and efficiency can be achieved by training intrinsically sparse neural networks from scratch. Existing sparse training methods usually strive to find the best sparse subnetwork possible in one single run, without involving any expensive dense or pre-training steps. For instance, dynamic sparse training (DST), is capable of reaching a competitive performance of dense training by iteratively evolving the sparse topology during the course of training. In this paper, we argue that it is better to allocate the limited resources to create multiple low-loss sparse subnetworks and superpose them into a stronger one, instead of allocating all resources entirely to find an individual subnetwork. To achieve this, two desiderata are required: (1) efficiently producing many low-loss subnetworks, the so-called cheap tickets, within one training process limited to the standard training time used in dense training; (2) effectively superposing these cheap tickets into one stronger subnetwork. To corroborate our conjecture, we present a novel sparse training approach, termed \textbf{Sup-tickets}, which can satisfy the above two desiderata concurrently in a single sparse-to-sparse training process. Across various modern architectures on CIFAR-10/100 and ImageNet, we show that Sup-tickets integrates seamlessly with the existing sparse training methods and demonstrates consistent performance improvement.
\vspace{-0.5em}
\end{abstract}

\begin{figure}[htbp]
\vskip 0.2in
\begin{center}
\centerline{\includegraphics[width=0.45\textwidth]{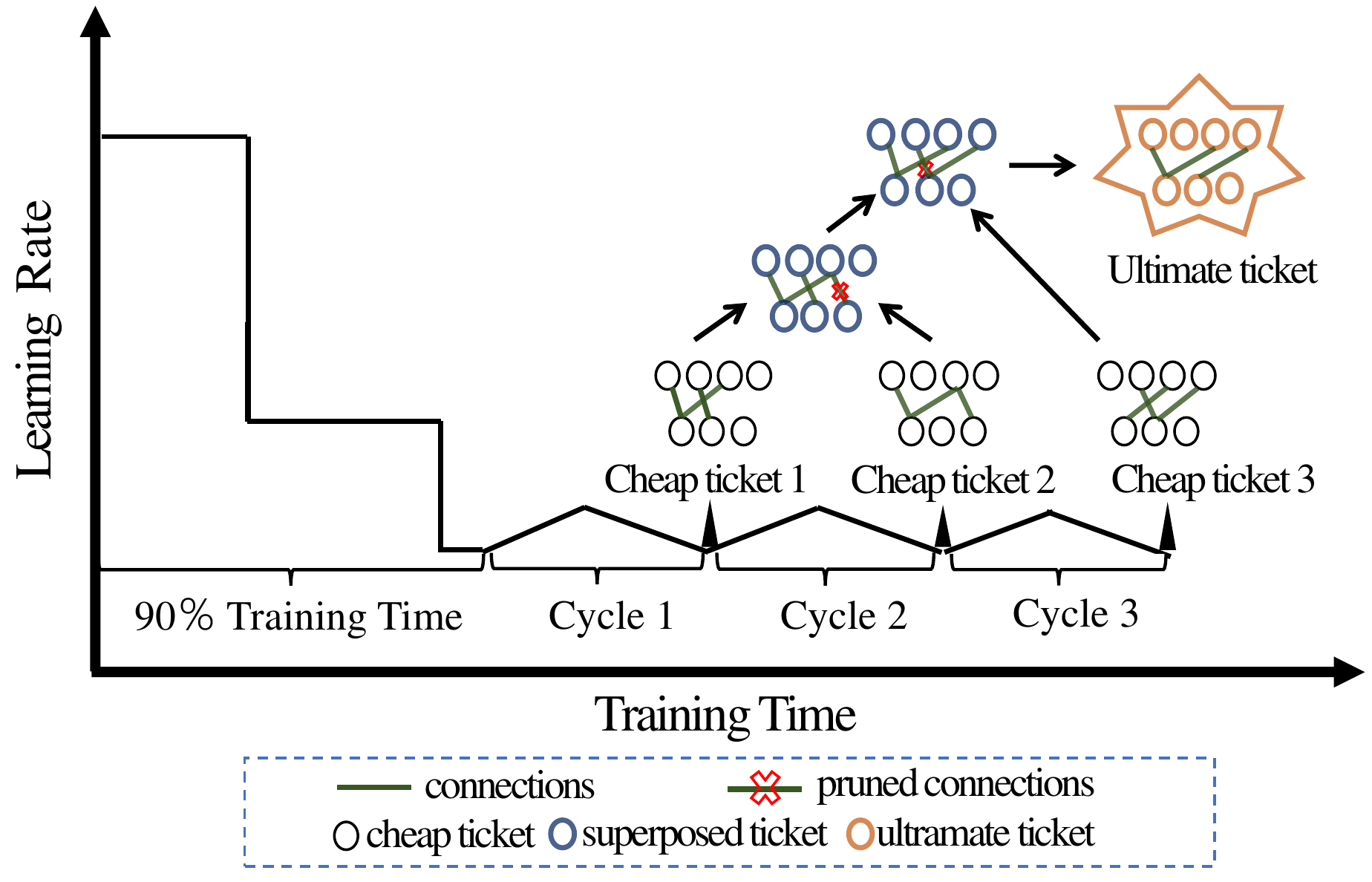}}

\caption{The schematic view of Sup-tickets. Multiple subnetworks (cheap tickets) are efficiently produced within the last 10\% of the training time and are superposed into one single subnetwork with boosting performance while maintaining the target sparsity. We term the ``ultimate ticket'' as the final subnetwork used for inference. }
\label{fig:Sup-tickets}
\end{center}
\vspace{-2em}
\end{figure}

\section{Introduction}
\vspace{-0.5em}

Over the past years, large-scale deep learning models with billions, even trillions of parameters have improved the state-of-the-art in nearly every downstream task~\citep{shoeybi2019megatron,brown2020language,radford2021learning,fedus2021switch}. The compelling results achieved by these large-scale models motivate researchers to pursue increasingly gigantic models without thinking too much about the limited resources of our planet. Fortunately, many prior techniques for neural network acceleration have already been proposed, which can effectively trim down the memory requirements and computational costs while retaining high accuracy~\citep{mozer1989using,han2015deep,gale2019state,molchanov2017variational}. 

Among them, sparse neural network training~\citep{mocanu2018scalable,evci2020rigging,bellec2018deep} stands out and receives growing attention recently due to its high efficiency in both the training and inference phases. Instead of inheriting well-performing sparse networks from a trained dense network, sparse training approaches typically start from a randomly initialized sparse network and only require training a subset of the corresponding dense network. Since this sparse-to-sparse training process does not involve any dense or pre-training steps, the memory requirements and the floating-point operations (FLOPs) are only a fraction of the traditional dense training. Nonetheless, naively training a sparse neural network from scratch leads to poor solutions in general compared with training a dense network~\citep{evci2019difficulty}. Dynamic sparse training (DST)~\citep{mocanu2018scalable} significantly improves the trainability of sparse networks by dynamically exploring new connectivities during training, while maintaining the fixed parameter count. Compared with methods that train with the fixed sparse connectivity~\citep{Mocanu2016xbm,lee2018snip}, DST substantially improves the expressibility of sparse networks, and thus leads to better generalization performance~\citep{liu2021we}. {However, the accuracy of extremely sparse subnetworks (e.g., at sparsity\footnote{The term sparsity refers to the proportion of the neural network's weights that are zero-valued.} 95\% or 90\%) usually remains below the full dense training under a regular training epoch number~\citep{evci2020rigging,liu2021neuroregeneration}. Enabling sparse training at extreme sparsities to match or even surpass the performance of dense training under a typical amount of training epochs will significantly 
benefit sparse training in practice.} 


Increasingly more evidence on sparse training~\citep{liu2021deep} and dense training~\citep{garipov2018loss,draxler2018essentially,fort2019large} reveal that many independent local optima exist in different low-loss basins of the loss landscape.  Inspired by these observations, we go one step further to pursue an approach that can boost the performance of sparse training by leveraging these widely-existing low-loss basins. Specifically, we propose  Superposing Tickets, or briefly~\textbf{Sup-tickets}, which could produce many subnetworks (cheap tickets) in one single run and then superposes all of them into one at the same sparsity. Doing so allows us to leverage the knowledge from various well-performing cheap tickets, while still maintaining the training and inference efficiency of sparse training.  Overall, we summarize our contributions below:



\begin{itemize}
    \item We propose Sup-tickets, a novel sparse training approach that produces and superposes many cheap yet well-performing subnetworks (cheap tickets) during one sparse-to-sparse training run. The ultimate superposed subnetwork achieves stronger results in predictive accuracy and uncertainty estimation while maintaining the target sparsity. 
    
    
    \item  {
Sup-tickets is a general and versatile performance booster for sparse training, which seamlessly integrates with other state-of-the-art sparse training methods. We conduct extensive experiments to evaluate our method. Across various popular architectures on CIFAR-10/100 and ImageNet, Sup-tickets improves the performance of various sparse training methods without extending the training time.}
    
    \item  {More impressively, in conjunction with the advanced sparse training methods -- GraNet~\citep{liu2021neuroregeneration}, Sup-tickets boosts the performance of sparse training over the dense training on CIFAR-10/100 at extreme sparsity levels around 90\% $\sim$ 95\%, enhancing the great potentials of sparse training in practice.}

\end{itemize}

\section{Related Work}

\subsection{Sparse Neural Network Training}
Sparse neural network training is a thriving topic. It aims to train initial sparse neural networks from scratch and chase competitive performance with their dense counterparts, while using only a fraction of resources of the latter. According to whether the sparse connectivity dynamically changes or not during training, sparse training usually can be divided into static sparse training (SST) and dynamic sparse training (DST).

\textbf{Static sparse training} represents a class of methods that train initial sparse neural networks with a fixed sparse connectivity pattern throughout training. While the sparse connectivity is static, the choices of the particular layer-wise sparsity (i.e., sparsity level of every single layer) can be diverse. The most naive approach is sparsifying each layer uniformly, i.e., uniform sparsity~\citep{gale2019state}.~\citet{Mocanu2016xbm} proposed a non-uniform sparsity method that can be applied in Restricted Boltzmann Machines (RBMs) and achieves better performance than dense RBMs. Some works explore 
the expander graph to train sparse CNNs and show comparable performance against the corresponding dense CNNs~\citep{prabhu2018deep,kepner2019radix}. Inspired by the graph theory, \textit{Erd{\H{o}}s-R{\'e}nyi} (ER)~\citep{mocanu2018scalable} and its CNNs variant \textit{Erd{\H{o}}s-R{\'e}nyi-Kernel} (ERK)~\citep{evci2020rigging} allocates lower sparsity to smaller layers, avoiding the layer collapse problem~\citep{tanaka2020pruning} and achieving stronger results than the uniform sparsity in general.

\textbf{Dynamic sparse training}, namely, trains initial sparse neural networks while dynamically adjusting the sparse connectivity pattern during training. DST was first introduced in Sparse Evolutionary Training (SET)~\citep{mocanu2018scalable} which initializes the sparse connectivity with a ER topology and periodically explores the parameter space via a prune-and-grow scheme during training. Following SET, weights redistribution is introduced to search for better layer-wise sparsity ratios while training~\citep{mostafa2019parameter,dettmers2019sparse}. The mainly-used pruning criterion of existing DST methods is magnitude pruning. The criterion used for weight regrowing varies from method to method. Gradient-based regrowth e.g., momentum~\citep{dettmers2019sparse} and gradient~\citep{evci2020rigging}, shows strong results in image classification, whereas random regrowth outperforms the former in language modeling~\citep{dietrich2021towards}. Follow-up works improve the accuracy by relaxing the constrained memory footprint ~\citep{jayakumar2020top,yuan2021mest,liu2021neuroregeneration,huang2022fedspa}. Very recently,~\citet{liu2021deep} proposed an efficient ensemble framework for sparse training-- FreeTickets. By directly ensembling the predictions of individual subnetworks, FreeTickets surpass the generalization performance of  the naive dense ensemble. Nevertheless, FreeTickets requires extending the training time to obtain multiple cheap subnetworks and performing multiple forward passes for inference, contrary to our pursuit of efficient training.

\subsection{Weight Averaging}

{Computing the convex combination of model weights usually leads to better robust performance~\cite{zhang2019lookahead,neyshabur2020being,wortsman2022model}. SWA~\citep{izmailov2018averaging} average weights along the same optimization trajectory with one single run. \cite{neyshabur2020being}, in contrast, merge models that start with the same initialization but are optimized independently.  Similarly, \cite{wortsman2022model} average models across many independent runs with various hyperparameters.  Different from these prior works that only study on dense networks, we explore for the first time how to produce and combine multiple \textit{sparse sub-networks}  into a stronger one while considering the importance of the connectivities.}





\section{Methodology}

\begin{algorithm*}[tb]
   \caption{Sup-tickets}
   \label{alg:example}
\begin{algorithmic}[1]
\REQUIRE Network $f(\bm{x}; \bm{\mathrm{\theta}})$,  superposed subnetwork $\widetilde{\bm{\mathrm{\theta}}}_\mathrm{s}$, target sparsity $S$, training time $T$, cycle length $C$, learning rate $\alpha$, pruning criterion $\Psi$,  growing criterion $\Phi$,  pruning rate for parameter exploration $p$.

\STATE $f(\bm{x}; \bm{\mathrm{\theta}}_s)$ $\gets$  $f(\bm{x}; \bm{\mathrm{\theta}}; S)$  \Comment{Sparsely initialize the network}
\FOR {$i \leftarrow 1$ \bf{to} $T$}  
\IF{$i \leq 90\% T$}   \Comment{Normal sparse training for the first 90\% of T}
\STATE $f(\bm{x}; \bm{\mathrm{\theta}}_s) \gets$ $SparseTraining(f(\bm{x}; \bm{\mathrm{\theta}}_s))$ 
\ELSE                  \Comment{Creating and superposing cheap tickets in the last 10\% of T}
\STATE $\alpha \leftarrow \alpha(i) $                                                                  \Comment{Calculate the cyclical learning rate using Eq.~\ref{eq:cyc_schedule}}
\STATE $f(\bm{x}; \bm{\mathrm{\theta}}_s) \gets$ $SparseTraining(f(\bm{x}; \bm{\mathrm{\theta}}_s); \alpha)$ 
\IF{$\bmod(i-90\%T, C) = 0$}     
    \STATE  $t \gets (i-90\%T)/C$              \Comment{Number of the created cheap tickets}
     
    \STATE 
$\widetilde{\bm{\mathrm{\theta}}}_\mathrm{s}^\mathrm{t} \gets \frac{(t-1) \cdot \widetilde{\bm{\mathrm{\theta}}}_\mathrm{s}^\mathrm{t-1} + \bm{\mathrm{\theta}}_\mathrm{s}^\mathrm{t}}{t}$  \Comment{  Ticket superposing using Eq.~\ref{eq:average} }
    \STATE 
    $ \widetilde{\bm{\mathrm{\theta}}}_\mathrm{s}^\mathrm{t} \gets MagnitudePruning(\widetilde{\bm{\mathrm{\theta}}}_\mathrm{s}^\mathrm{t})$ 
    \Comment{Prune the superposed ticket to the target sparsity $S$}
    \STATE 
    $\bm{\mathrm{\theta}}_{\mathrm{s}}^\prime  \gets \Psi(\bm{\mathrm{\theta}}_\mathrm{s},~p)$  \Comment{Parameter exploration using Eq.~\ref{eq:prune} and Eq.~\ref{eq:regrow}}  
    \STATE  $\bm{\mathrm{\theta}}_\mathrm{s} \gets  \bm{\mathrm{\theta}}_{\mathrm{s}}^\prime \cup     \Phi(\bm{\mathrm{\theta}}_{i \notin \bm{\mathrm{\theta}}_\mathrm{{s}^\prime}},~p)$   
\ENDIF
\ENDIF
\ENDFOR 
\STATE Return $\widetilde{\bm{\mathrm{\theta}}}_\mathrm{s}$  \COMMENT{The ultimate ticket for test}



    

\end{algorithmic}
\end{algorithm*}

In this section, we introduce a new approach for sparse training, which could combines the benefits of multiple cheap tickets, without extra training time and multiple forward passes for inference\citep{garipov2018loss,liu2021deep}. We first introduce the basic training scheme of sparse training in Section~\ref{sec:sparse_training} and then describe our proposed Sup-tickets approach in detail in Section~\ref{sec:sup_ticket}.

\subsection{Prior Sparse Training Art}
\vspace{-0.5em}
\label{sec:sparse_training}

Following~\citet{liu2021we,liu2021deep}, we denote a sparse neural network as $f(\bm{x}; \bm{\mathrm{\theta}}_\mathrm{s})$. $\bm{\mathrm{\theta}}_\mathrm{s}$ refers to a subset of the full network parameters $\bm{\mathrm{\theta}}$ at a sparsity level of ${(1 - \frac{\|\bm{\mathrm{\theta}}_\mathrm{s}\|_0}{\|\bm{\mathrm{\theta}}\|_0}})$, where $\|\cdot\|_0$ is the $\ell_0$-norm. Sparse training typically initializes the network in a random fashion where the connections between two adjacent layers are sparsely and randomly connected, based on a pre-defined uniform or non-uniform layer-wise sparsity ratio\footnote{See~\citet{liu2022the} for the most common types of sparse initialization.}. In the i.i.d. classification setting with data $\{(x_i, y_i) \}_{i=1}^{\mathrm{N}}$, the goal of sparse training is to solve the following optimization problem: $\hat{\bm{\mathrm{\theta}}_\mathrm{s}} = \argmin_{\bm{\mathrm{\theta}}_\mathrm{s}} \sum_{i=1}^{\mathrm{N}} \mathcal{L}(f(x_i;\bm{\mathrm{\theta}}_\mathrm{s}),y_i)$, where $\mathcal{L}$ is the loss function.  

SST keeps the sparse connectivity of the sparse network fixed after initialization. DST, on the other hand, dynamically adjusts the sparse connectivity via parameter exploration during training while sticking to a fixed sparsity level. The most widely used method for parameter exploration is the prune-and-grow scheme, i.e., pruning $p\%$ the least important parameters from the current subnetwork followed by a fraction $p\%$ of weight growing. Formally, the parameter exploration can be written as the following two steps:
\begin{equation}
    \bm{\mathrm{\theta}}_{\mathrm{s}}^\prime = \Psi(\bm{\mathrm{\theta}}_\mathrm{s},~p),
    \label{eq:prune}
\end{equation}
\vspace{-1.5em}
\begin{equation}
    \bm{\mathrm{\theta}}_\mathrm{s} = \bm{\mathrm{\theta}}_{\mathrm{s}}^\prime
    \cup \Phi(\bm{\mathrm{\theta}}_{i \notin \bm{\mathrm{\theta}}_\mathrm{{s}}^\prime},~p)
    \label{eq:regrow}
\end{equation}
where $\Psi$ and $\Phi$ are the specific pruning and growing criterion respectively. The choices of $\Psi$ and $\Phi$ differ from sparse training method to another. Besides the sparse structures, in the most sparse training literature~\citep{dettmers2019sparse,evci2020rigging,mostafa2019parameter,liu2021neuroregeneration}, it is usually a safe choice to keep the other training configurations, such as optimizers, hyperparameters, and learning rate schedules, the same as the normal dense training.  At the end of the training, sparse training can converge to a well-performing sparse subnetwork whose memory requirements, training, and inference FLOPs are only a fraction of the dense training.


\subsection{Sup-tickets}
\vspace{-0.5em}
\label{sec:sup_ticket}
Existing sparse training methods allocate all the limited resources to find the best sparse neural network possible. While low-loss subnetworks widely exist in the loss landscape of sparse neural network optimization~\citep{liu2021sparse}, no prior works have ever explored how to find and leverage these handy cheap tickets to boost the performance of sparse training without extending training steps. In this section, we present Sub-tickets to close this research gap, as illustrated in Figure~\ref{fig:Sup-tickets}.

To achieve the above-mentioned ultimate goal, we need to satisfy the following two desiderata in one sparse-to-sparse training run: 
\begin{enumerate}
    \item \textbf{Creating cheap tickets}: Creating multiple cheap but well-performing subnetworks with one single run under a regular training time. We name such efficiently produced subnetworks as ``cheap tickets''.
    \item \textbf{Superposing tickets}: Superposing these subnetworks into one subnetwork at the same sparsity to avoid performing multiple forward passes for the prediction. We term the ``ultimate ticket'' as the final subnetwork used for inference.
\end{enumerate}
These two desiderata strictly follow the sparsity constraint of sparse training and thus maintain the training/inference efficiency of sparse training.

\vspace{-1em}
\subsubsection{Creating Cheap Tickets}
\begin{figure}[h]
\centering
    \subfigure{}{
        \includegraphics[width=0.45\textwidth]{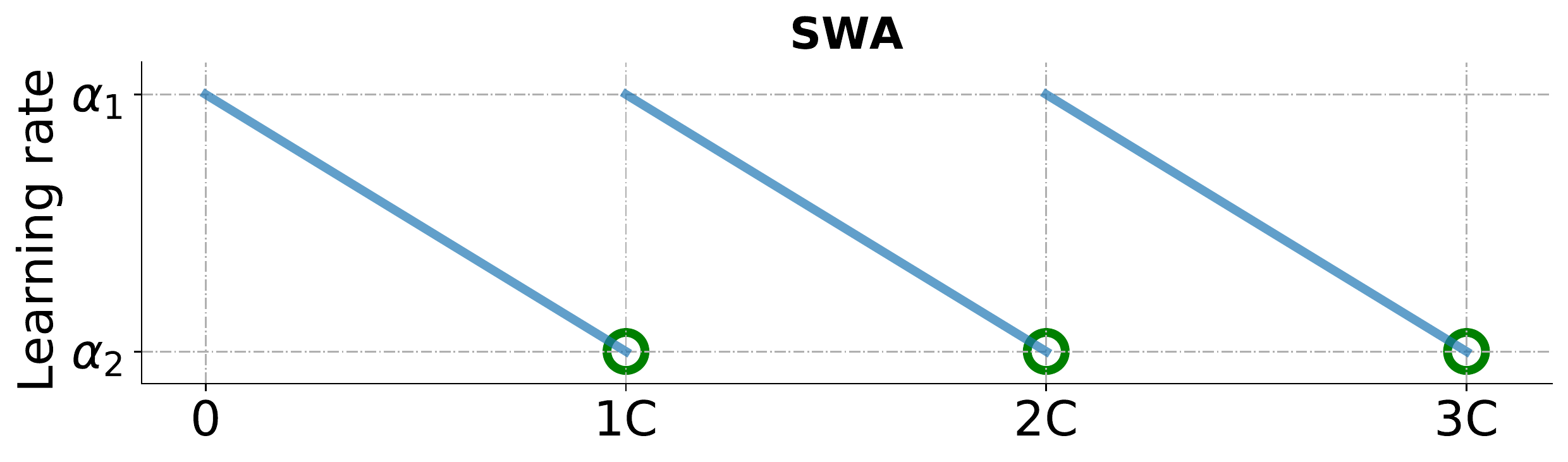}
    }
    \subfigure{}{
        \includegraphics[width=0.45\textwidth]{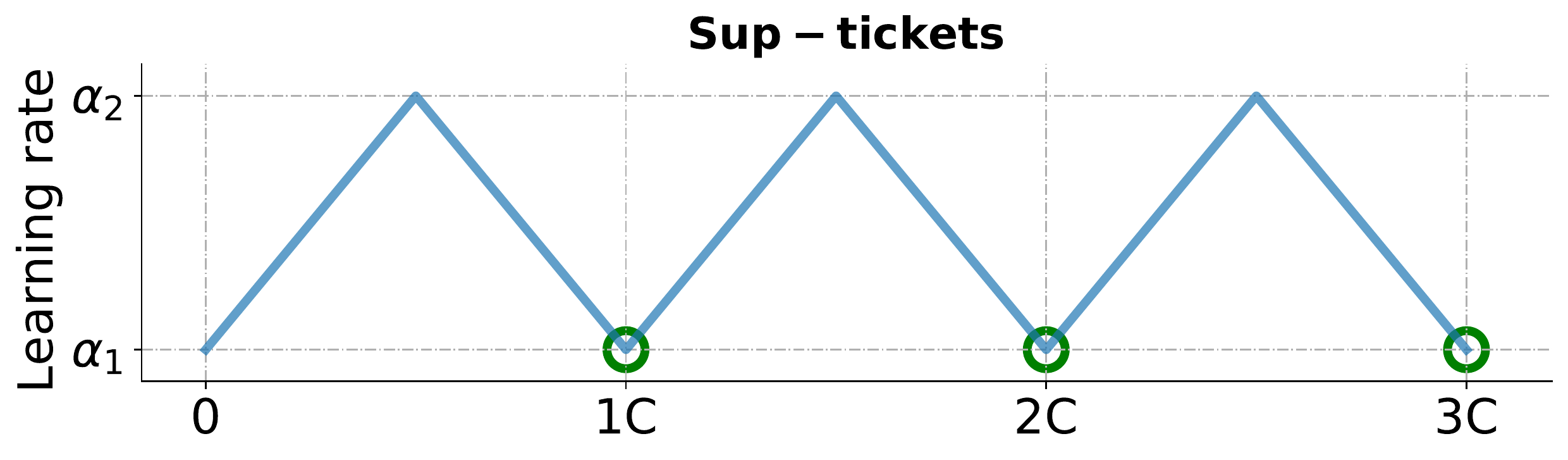}
    }

\caption{\textbf{Top:} cyclical learning rate schedule of~\citet{garipov2018loss}. \textbf{Bottom:} cyclical learning rate schedule of Sup-tickets. Cheap tickets are collected at the end of each learning rate schedule cycle (green circles in the figure).}
\label{fig:learning rate reschedual}
\vspace{-1em}
\end{figure}

During the last 10\% of the training time,  we cyclically explore the current sparse connectivity  and restart the learning rate to visit multiple low-loss sub-space basins.
More concretely, in each cycle, we first significantly change the connectivity of the current subnetwork by performing the parameter exploration once with Eq.~\ref{eq:prune} $\&$~\ref{eq:regrow}. For simplicity, we inherit the pruning and growing methods used in the sparse training methods that Sup-tickets combines with. After parameter exploration, we leverage the cyclical learning rate to force the current subnetwork to escape the local minima. Inspired by~\citet{garipov2018loss,izmailov2018averaging}, we adopt the learning rate schedule scheme as:

\vspace{-1em}
\begin{equation}
\label{eq:cyc_schedule}
\resizebox{.90\hsize}{!}{$
        \alpha(i)\!= \left\{ 
        \begin{array}{ll}
        (1 - 2 t(i)) \alpha_1 + 2 t(i) \alpha_2 &  0 < t(i) \le \frac{1}{2} \\
        (2 - 2 t(i)) \alpha_2 + (2 t(i) - 1) \alpha_1 & \frac{1}{2} < t(i) \le 1 
        \end{array} 
    \right.$}  
\end{equation}

where $\alpha(i)$ is the cyclical learning rate ranging from $\alpha_1$ to $\alpha_2$; $i$ is the training iteration for one mini-batch data; $t(i) = \frac{1}{C}(\bmod(i - 1, C) + 1)$; $C$ is the cycle length. We modify the cyclical learning rate schedule used in SWA~\citep{izmailov2018averaging} to prevent the aggressive rise of the learning rate. Specifically, we adopt the triangle-like schedule as shown in Figure~\ref{fig:learning rate reschedual}-bottom. In such a way, the learning rate could seamlessly transition from the normal training stage to the superposing stage. At the end of each cycle, we can obtain one cheap ticket from the current basin with 
diverse and meaningful representation.  

The combination of cyclical learning rate schedule and parameter exploration is also used in FreeTickets~\citep{liu2021deep}, but we have several distinctions to make it compiled with the requirements of sparse training. The cycle duration of FreeTickets is set as 100 epochs to guarantee the consistent strong performance of each subnetwork as they try to achieve comparable performance with the dense ensemble. However, such a long duration of cycle conflicts with the goal of sparse training. In particular, we reduce the cycle duration to 2 epochs for ImageNet, 8 epochs for CIFAR-10/100 and only use the final 10\% of the training time to generate cheap tickets. In this case, the overall training time is the same as training a single sparse network.

\begin{figure*}[htbp]
\centering
    \subfigure{}{
        \includegraphics[width=01\textwidth]{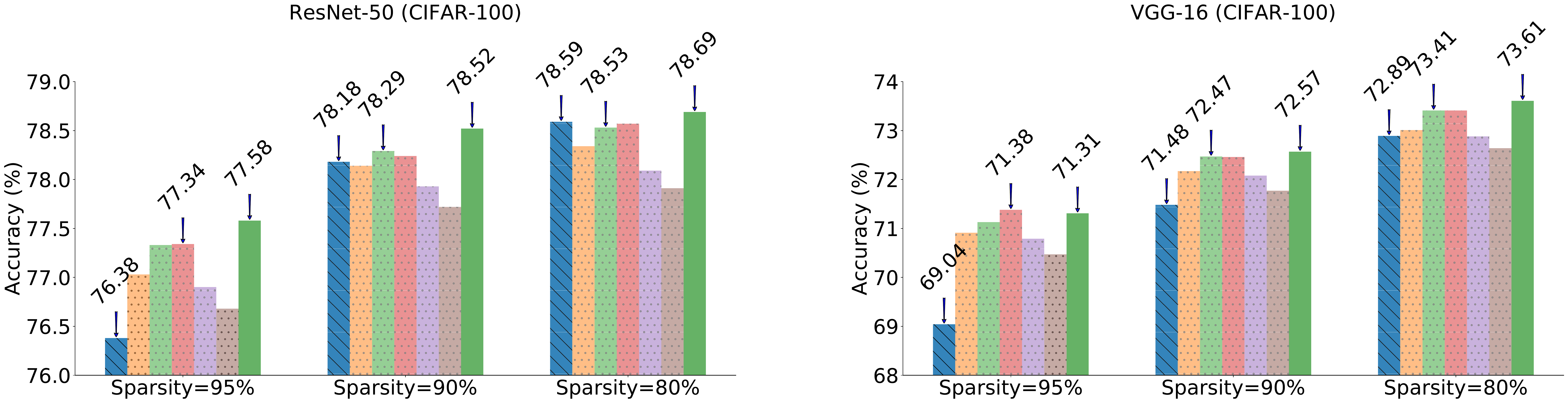}
    }
    \vspace{-1em}
    \subfigure{}{
        \includegraphics[width=0.8\textwidth]{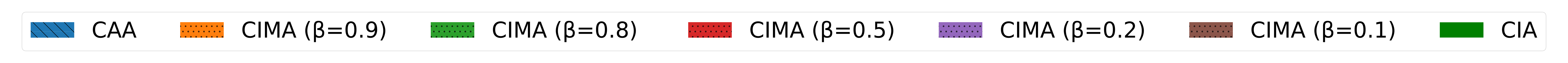}
    }

\caption{Comparisons of various averaging methods. We combine CIA, CAA, and CIMA  with RigL and report the test accuracy of the ultimate tickets. For CIMA, we vary the exponential decay rates $\beta \in [0.9, 0.8, 0.5, 0.2, 0.1]$.}
    \label{Average_methods}

\end{figure*}

\subsubsection{Superposing Tickets}\label{S_tickets}

\vspace{-0.2cm}
Superposing multiple sparse networks is more complex than superposing multiple dense networks~\citep{cheung2019superposition,izmailov2018averaging}. Naively selecting all the weights that are activated in all cheap tickets will significantly increase the parameter count, as different subnetworks have different connectivities. To solve this task, we propose to perform \underline{weight averaging followed by weight pruning}. More concretely, assuming we collect $\mathrm{M}$  cheap tickets $\{\bm{\mathrm{\theta}}_\mathrm{s}^1, \bm{\mathrm{\theta}}_\mathrm{s}^2, ... , \bm{\mathrm{\theta}}_\mathrm{s}^\mathrm{M}\}$ at the end of training, we consider the following three ways to average them.

\textbf{Connection Independent Averaging (CIA).} The ultimate subnetwork averaged by CIA is given as: $\widetilde{\bm{\mathrm{\theta}}}_\mathrm{s^{\prime}} = \frac{1}{\mathrm{M}} \sum_{\mathrm{i=1}}^{\mathrm{M}} \bm{\mathrm{\theta}}_\mathrm{s}^\mathrm{i}$, where $\mathrm{M}$ is the total number of cheap tickets. CIA simply averages weights across all the cheap tickets without considering whether the connection is activated or not in each cheap ticket. CIA tends to preserve the connections that are activated in the majority of the cheap tickets whereas the ones that are occasionally activated in one or two cheap tickets are likely to have small magnitude after averaging by $\mathrm{M}$, unless they have extremely large values.

\textbf{Connection Aware Averaging (CAA).}   The ultimate subnetwork averaged by  CAA  is given as: $\widetilde{\bm{\mathrm{\theta}}}_\mathrm{s} = \frac{1}{\mathrm{N(k, j)}} \sum_{\mathrm{i=1}}^{\mathrm{M}} \bm{\mathrm{\theta}}_\mathrm{s}^\mathrm{i}$, where $\mathrm{N(k, j)}$ is the number of times the connection $\mathrm{\theta(k, j)}$ is activated across all the cheap tickets; $k$ is the $k^{th}$ neuron in the previous layer and  $j$ is the  $j^{th}$ neuron in this layer. Thus, we have $\mathrm{N(k, j)} \leq \mathrm{M}$. Compared with CIA, CAA pays more attention to the occasionally activated connections that are only existing in the minority of cheap tickets.

\textbf{Connection Independent Moving Averaging (CIMA).}
Motivated by the widely-used moving average technique~\citep{rumelhart1986learning,kingma2014adam,karras2017progressive}, we sequentially apply the popular moving averages over the cheap tickets obtained at each cycle. The averaged subnetwork over the first $t$ cheap tickets is given as:  $\widetilde{\bm{\mathrm{\theta}}}_\mathrm{s}^\mathrm{t} = \beta \widetilde{\bm{\mathrm{\theta}}}_\mathrm{s}^\mathrm{t-1}  + (1-\beta) \bm{\mathrm{\theta}}_\mathrm{s}^\mathrm{t}$.  $\beta$ controls the exponential decay rates. Larger $\beta$ will put more emphasis on the cheap tickets collected in the early time.

Note that the sparsity of the averaged subnetwork is likely larger than the target sparsity level. To maintain the same sparsity as the original subnetwork, we utilize magnitude weight pruning to remove the weights with the smallest magnitude after every averaging step.

\subsection{Memory and Computation Overhead}

Instead of saving $\mathrm{M}$ individual cheap tickets and average them, we apply a similar operation as used in CIMA to save the extra memory required by CIA and CAA during training. The averaged subnetwork over the first $t$ cheap tickets is given as:
\begin{equation}
\widetilde{\bm{\mathrm{\theta}}}_\mathrm{s}^\mathrm{t} = \frac{(t-1) \cdot \widetilde{\bm{\mathrm{\theta}}}_\mathrm{s}^\mathrm{t-1} + \bm{\mathrm{\theta}}_\mathrm{s}^\mathrm{t}}{t}
\label{eq:average}
\end{equation}
This operation allows us to accomplish the average operation by maintaining only one extra copy of the averaged weights, instead of saving $\mathrm{M}$ subnetworks.

Moreover, as we mentioned, we use the final 10\% of the training time to create cheap tickets, and thus the training time of Sub-tickets is the same as the standard sparse training. Since we only need to perform Eq.~\ref{eq:average} for ($\mathrm{M}-1$) times, the extra computation cost of averaging is negligible compared with the total training costs. Overall, we can conclude that the training cost of Sub-tickets is approximately the same as training a single sparse network.

\section{Experiments}

Sub-tickets is a universal idea that can be  straightforwardly applied to any types of sparse training methods. To verify the effectiveness of Sup-tickets, we apply it to various sparse training methods, including 3 DST methods: SET, RigL~\citep{evci2020rigging}, and GraNet~\citep{liu2021neuroregeneration}; one SST method: ERK~\citep{evci2020rigging}; and one pruning at initialization approach: SNIP~\citep{lee2018snip}.

\begin{table*}[htbp]
\centering
\caption{Test accuracy (\%) of sparse VGG-16 on CIFAR-10/100. All the results are averaged from three random runs. In each setting, the best results are marked in bold.}
\vspace{-1em}
\label{table:VGG16}
\resizebox{1.0\textwidth}{!}{
\begin{tabular}{lccc ccc}
\cmidrule[\heavyrulewidth](lr){1-7}

 \textbf{Dataset}     & \multicolumn{3}{c}{CIFAR-10} & \multicolumn{3}{c}{CIFAR-100}  \\ 
 \cmidrule(lr){1-1}
\cmidrule(lr){2-4}
\cmidrule(lr){5-7}
 
 \textbf{VGG-16}~(Dense) 
& 93.91$\pm$0.26  & - & - 
& 73.61$\pm$0.45  & - & - 
\\
 \cmidrule(lr){1-1}
\cmidrule(lr){2-4}
\cmidrule(lr){5-7}

Sparsity     & 95\%      & 90\%     & 80\%     
     &  95\%      & 90\%     & 80\%         \\ 
     
 \cmidrule(lr){1-1}
\cmidrule(lr){2-4}
\cmidrule(lr){5-7}

SET~\citep{mocanu2018scalable}
& 92.96$\pm$0.18 & 93.54$\pm$0.23   & 93.56$\pm$0.04
& 70.10$\pm$0.33 & 71.50$\pm$0.23  & 72.38$\pm$0.08

\\
SET+Sup-tickets (ours)
&\textbf{93.22$\pm$0.09}  & \textbf{93.63$\pm$0.05} & \textbf{93.80$\pm$0.13} &
\textbf{71.18$\pm$0.29}  & \textbf{71.99$\pm$0.27} & \textbf{73.02$\pm$0.32}
\\
\cmidrule(lr){1-1}
\cmidrule(lr){2-4}
\cmidrule(lr){5-7}
RigL~\citep{evci2020rigging}
& 92.70$\pm$0.08 & 93.48$\pm$0.16   &93.60$\pm$0.14
& 70.65$\pm$0.16   & 72.20$\pm$0.09  & 72.63$\pm$0.23 
\\
RigL+Sup-tickets (ours)
&\textbf{93.20$\pm$0.13}  &  \textbf{93.81$\pm$0.11}  & \textbf{93.85$\pm$0.25}
&\textbf{71.31$\pm$0.21} & \textbf{72.57$\pm$0.29} &  {\textbf{73.61$\pm$0.11}}
\\
\cmidrule(lr){1-1}
\cmidrule(lr){2-4}
\cmidrule(lr){5-7}
GraNet \citep{liu2021neuroregeneration}  
& 93.87$\pm$0.19 & 93.83$\pm$0.30 & 93.77$\pm$0.18
&72.91$\pm$0.39 & 73.48$\pm$0.17 & 73.36$\pm$0.14

\\
GraNet+Sup-tickets (ours)
& {\textbf{94.10$\pm$0.06}} &  {\textbf{94.13$\pm$0.12}} &  {\textbf{94.24$\pm$0.05}}
&  {\textbf{73.61$\pm$0.24}}&  {\textbf{73.87$\pm$0.26}} &  {\textbf{73.95$\pm$0.30}}

\\
\cmidrule[\heavyrulewidth](lr){1-7}
\vspace{-2em}
\end{tabular}}
\end{table*}

\begin{table*}[htbp]
\centering
\caption{Test accuracy (\%) of sparse ResNet-50 on CIFAR-10/100. All the results are averaged from three runs. In each setting, the best results are marked in bold.}
\label{table:RN50_CIFAR}
\resizebox{1.0\textwidth}{!}{
\begin{tabular}{lccc ccc}
\cmidrule[\heavyrulewidth](lr){1-7}

 \textbf{Dataset}     & \multicolumn{3}{c}{CIFAR-10} & \multicolumn{3}{c}{CIFAR-100}  \\ 
\cmidrule(lr){1-1}
\cmidrule(lr){2-4}
\cmidrule(lr){5-7}
\textbf{ResNet-50}~(Dense) 
& 94.88$\pm$0.11 & - & -  
& 78.00$\pm$0.40  & - & - 
\\
\cmidrule(lr){1-1}
\cmidrule(lr){2-4}
\cmidrule(lr){5-7}

Sparsity     & 95\%      & 90\%     & 80\%     
     &  95\%      & 90\%     & 80\%         \\ 
     
\cmidrule(lr){1-1}
\cmidrule(lr){2-4}
\cmidrule(lr){5-7}

SNIP~\citep{lee2018snip} 
& 94.01$\pm$0.28 & 94.81$\pm$0.36 & 94.91$\pm$0.16
& 41.25$\pm$1.10  & 68.79$\pm$1.16 & 75.29$\pm$1.28
\\

SNIP+Sup-tickets (ours)
& \textbf{94.33$\pm$0.09} & \textbf{95.05$\pm$0.22} & \textbf{95.21$\pm$0.09}
& \textbf{65.56$\pm$1.15} & \textbf{76.34$\pm$0.27} & \textbf{77.43$\pm$0.53}

\\
\cmidrule(lr){1-1}
\cmidrule(lr){2-4}
\cmidrule(lr){5-7}

ERK~\citep{evci2020rigging} 
& 93.44$\pm$0.22 & 94.41$\pm$0.13 & 94.85$\pm$0.21
& 74.49$\pm$0.30 &  76.36$\pm$0.22 & 77.41$\pm$0.08
\\

ERK+Sup-tickets (ours)
& \textbf{93.92$\pm$0.04} & \textbf{94.80$\pm$0.06} & \textbf{95.11$\pm$0.27}
& \textbf{75.75$\pm$0.28} & \textbf{76.82$\pm$0.08} & \textbf{77.85$\pm$0.42}
\\
\cmidrule(lr){1-1}
\cmidrule(lr){2-4}
\cmidrule(lr){5-7}

SET~\citep{mocanu2018scalable}
&94.49$\pm$0.11   & 94.73$\pm$0.27   & 94.74$\pm$0.17
& 76.59$\pm$0.54  & 77.79$\pm$0.27   & \textbf{78.45$\pm$0.50}  
\\

SET+Sup-tickets (ours)
&\textbf{94.81$\pm$0.05}   & \textbf{94.87$\pm$0.03}  & \textbf{94.90$\pm$0.27}
&\textbf{76.68$\pm$0.38}  & \textbf{77.89$\pm$0.45}   &  {78.35$\pm$0.18}
\\

\cmidrule(lr){1-1}
\cmidrule(lr){2-4}
\cmidrule(lr){5-7}
RigL~\citep{evci2020rigging}
& 94.59$\pm$0.19 & 94.70$\pm$0.17   & 94.70$\pm$0.07
& 76.96$\pm$0.39 & 77.95$\pm$0.36   & {78.19$\pm$0.51}
\\
RigL+Sup-tickets (ours)
& \textbf{94.65$\pm$0.11}  & \textbf{94.82$\pm$0.13} &  \textbf{94.81$\pm$0.15}
&\textbf{77.58$\pm$0.47}  &  {\textbf{78.52$\pm$0.39}}  &    {\textbf{78.69$\pm$0.30}}
\\
\cmidrule(lr){1-1}
\cmidrule(lr){2-4}
\cmidrule(lr){5-7}

GraNet \citep{liu2021neuroregeneration}  
&94.70$\pm$0.23 &  94.95$\pm$0.09 & 94.86$\pm$0.24
&77.47$\pm$0.22 &  {78.25$ \pm$0.51} & {78.80$\pm$0.46}
\\
GraNet+Sup-tickets (ours)
& {\textbf{94.89$\pm$0.15}} &  {\textbf{95.08$\pm$0.08}} &  {\textbf{94.94$\pm$0.03}}
&\textbf{77.70$\pm$0.47} &  {\textbf{78.37$\pm$0.53}} &  {\textbf{78.95$\pm$0.33}}
\\

\cmidrule[\heavyrulewidth](lr){1-7}
\end{tabular}}
\end{table*}

\begin{table*}[htbp]
\centering
\caption{Test accuracy (\%) of sparse ResNet-50 on ImageNet. The training FLOPs of sparse training methods are normalized with the FLOPs used to train a dense dense model. In each setting, the best results are marked in bold.}
\vspace{0.1cm}
\label{table:classification_ImageNet}
\resizebox{1.0\textwidth}{!}{
\begin{tabular}{l  ccc  ccc}
\cmidrule[\heavyrulewidth](lr){1-7}
\textbf{Method} & Top-1 & FLOPs & FLOPs & TOP-1 & FLOPs & FLOPs \\
& Accuracy & (Train) & (Test) & Accuracy & (Train) & (Test)
\\
\cmidrule(lr){1-1}
\cmidrule(lr){2-4}
\cmidrule(lr){5-7}
\textbf{ResNet-50}~(Dense)  & 76.8$\pm$0.09 & 1x (3.2e18) & 1x (8.2e9) & 76.8$\pm$0.09 & 1x (3.2e18) & 1x (8.2e9)
\\
\cmidrule(lr){1-1}
\cmidrule(lr){2-4}
\cmidrule(lr){5-7}
Sparsity & \multicolumn{3}{c}{80\%} & \multicolumn{3}{c}{90\%}
\\
\cmidrule(lr){1-1}
\cmidrule(lr){2-4}
\cmidrule(lr){5-7}
Static sparse training (ERK) & 72.1$\pm$0.04 & 0.42$\times$ & 0.42$\times$ & 67.7$\pm$0.12 & 0.24$\times$ & 0.24$\times$ 
\\
Small-Dense & 72.1$\pm$0.06 & 0.23$\times$ & 0.23$\times$ & 67.2$\pm$0.12 & 0.10$\times$ & 0.10$\times$ 
\\
SNIP~\citep{lee2018snip} & 72.0$\pm$0.06 & 0.23$\times$ & 0.23$\times$ & 67.2$\pm$0.12 & 0.10$\times$ & 0.10$\times$ 
\\
SET~\citep{mocanu2018scalable} & 72.9$\pm${0.39} & 0.23$\times$ & 0.23$\times$ & 69.6$\pm${0.23} & 0.10$\times$ & 0.10$\times$ 
\\
DSR~\citep{mostafa2019parameter} & 73.3 & 0.40$\times$ & 0.40$\times$ & 71.6 & 0.30$\times$ & 0.30$\times$ 
\\
SNFS~\citep{dettmers2019sparse} & 75.2$\pm${0.11} & 0.61$\times$ & 0.42$\times$ & 72.9$\pm${0.06} & 0.50$\times$ & 0.24$\times$
\\
\cmidrule(lr){1-1}
\cmidrule(lr){2-4}
\cmidrule(lr){5-7}
RigL~\citep{evci2020rigging} & 75.1$\pm$0.05 & 0.42$\times$ & 0.42$\times$ & 73.0$\pm$0.04 & 0.25$\times$ & 0.24$\times$ 
\\
RigL+Sup-tickets (ours) & \textbf{76.0} & 0.42$\times$ & 0.42$\times$ & \textbf{74.0} & 0.25$\times$ & 0.24$\times$ 
\\
\cmidrule(lr){1-1}
\cmidrule(lr){2-4}
\cmidrule(lr){5-7}
GraNet~\citep{liu2021neuroregeneration} & 75.9 & 0.37$\times$  & 0.35$\times$
&  74.4 & 0.25$\times$ & 0.20$\times$
\\
GraNet+Sup-tickets (ours)  & {\textbf{76.2}} & 0.37$\times$  & 0.35$\times$
&  {\textbf{74.6}} & 0.25$\times$ & 0.20$\times$
\\
\cmidrule[\heavyrulewidth](lr){1-7}
\end{tabular}}
\vspace{-0.5cm}
\end{table*}

\subsection{Experimental Setups}

The experiments are conducted across various architectures on three popular datasets CIFAR-10/100 and ImageNet. For CIFAR-10/100, we choose models VGG-16~\citep{simonyan2014very}, Wide ResNet28-10~\citep{zagoruyko2016wide} and ResNet-50~\citep{he2016deep}. The models are trained for 250 epochs, optimized by momentum SGD with a learning rate of 0.1, which decayed by 10x at the half and three-quarters of the training stage. The cycle length is chosen as 8 epochs, so that we can obtain 3 cheap tickets in 24 epochs. The model used for ImageNet is ResNet-50, which is trained for 100 epochs, optimized by momentum SGD with a learning rate of 0.1 decaying by 10x at 30, 60, and 85 epoch. The cycle length of ImageNet is 2 epochs, so we obtain 4 cheap tickets in the last 8 epochs. The implementation details are reported in Appendix~\ref{sec:implementation}.

\subsection{Comparisons among CIA, CAA, and CIMA}

We first conduct a comparison among CIA, CAA, and CIMA on CIFAR-100 and report the results in Figure~\ref{Average_methods}. We can see that CIA consistently outperforms the other two methods at various sparsity levels. CAA is the worst-performing method, especially at the extreme sparsity 95\%. With tuned $\beta=0.8$, CIMA can approach the performance achieved by CIA. The better performance achieved by CIA over CAA indicates that the occasionally activated connections are likely unimportant. CIA pays more attention to the connections that exist in the majority of the cheap tickets, which can eliminate the unimportant connections that are activated occasionally. Therefore, due to the superior performance consistently achieved by CIA, we choose CIA as our averaging method in the following sections.

\subsection{Evaluation of Sup-tickets}
\label{sec:experiments sup-tickets}
\textbf{CIFAR-10/100.} In this section, we provide an experimental comparison of Sup-tickets to a variety of sparse training techniques. The results of CIFAR-10/100 with VGG-16 and ResNet-50 are shown in Table~\ref{table:VGG16} $\&$~\ref{table:RN50_CIFAR} respectively, and the results of Wide ResNet28-10 are shared in Appendix~\ref{sec:WRN2810} due to the limited space. Overall, we clearly see that our approach could benefit sparse training across all studied architectures. Simple as it looks, Sup-tickets improves the
performance of various dynamic sparse training methods in 63 out of 66 cases. It seems Sup-tickets performs better with VGG-16 than the other two architectures, with up to 0.5\% and 1.08\% accuracy increase on CIFAR-10 and CIFAR-100, respectively. We also find that the performance improvement on CIFAR-100 is larger than the one on CIFAR-10, which makes sense since CIFAR-100 is less saturated and thus has a larger improvement space. More importantly, our approach combined with the state-of-the-art DST method -- GraNet, outperforms the dense networks with only about 5\% at most 10\% parameters with all architectures, as reported in Table~\ref{table:GraNet Dense}. All these results highlight that Sup-tickets is a strong and universal performance booster for sparse training.

\begin{table}[!h]
    \centering
    \caption{{\small Performance comparison between  GraNet+Sup-tickets and dense network. Results that are better than the corresponding dense networks are marked in bold. WRN28-10 refers to Wide ResNet28-10. GraNet+Sup-tickets outperforms dense network in most cases.}}

    \label{table:GraNet Dense}
    \resizebox{0.48\textwidth}{!}{
    \begin{tabular}{lccccc}
        \hline
        \multirow{2}{*}{Dataset}  &  \multirow{2}{*}{Network} & \multirow{2}{*}{Dense}  & \multicolumn{3}{c}{ GraNet+Sup-tickets} \\
        \cmidrule(lr){4-6}
        & &   &   95\% sparsity & 90\% sparsity & 80\% sparsity   \\
        \cmidrule(lr){1-1}
        \cmidrule(lr){2-2}
        \cmidrule(lr){3-3}
        \cmidrule(lr){4-6}
        
        \multirow{3}{*}{CIFAR-10}  & VGG-16 & 93.91$\pm$0.26  &\textbf{{94.10$\pm$0.06}} & \textbf{{94.13$\pm$0.12}} & \textbf{{94.24$\pm$0.05}}\\ 
        & ResNet-50 & 94.88$\pm$0.11 &\textbf{{94.89$\pm$0.15}} & \textbf{{95.08$\pm$0.08}} & \textbf{{94.94$\pm$0.03}} \\
        & WRN28-10 &  96.00$\pm$0.13 & \textbf{{96.03$\pm$0.11}} & \textbf{{96.13$\pm$0.07}} & \textbf{96.08$\pm$0.04} \\

        \cmidrule(lr){1-1}
        \cmidrule(lr){2-2}
        \cmidrule(lr){3-3}
        \cmidrule(lr){4-6}
  
       \multirow{3}{*}{CIFAR-100} & VGG-16 & 73.61$\pm$0.45  & \textbf{{73.61$\pm$0.24}}& \textbf{{73.87$\pm$0.26}} & \textbf{{73.95$\pm$0.30}}\\
        & ResNet-50 & 78.00$\pm$0.40  &{77.70$\pm$0.47} & \textbf{{78.37$\pm$0.53}} & \textbf{{78.95$\pm$0.33}}\\ 
        & WRN28-10 & 81.09$\pm$0.19  &{80.65$\pm$0.06} & \textbf{{81.20$\pm$0.09}} & \textbf{{81.42$\pm$0.18}} \\ 
        
        \bottomrule
    \end{tabular}}

    \label{tab:performance_deberta}

\end{table}

\textbf{ImageNet.} For ImageNet, we apply Sup-tickets to RigL and GraNet and compare them with the existing sparse training methods. The results are reported the in Table~\ref{table:classification_ImageNet}. Again, we improve the performance of GraNet and RigL at both 80\% sparsity and 90\% sparsity without an extra parameter budget. Especially on RigL, our approach improves the test accuracy by 0.9\% and 1.0\% at sparsity 80\% and 90\%, respectively.  Besides, we compare the Sup-tickets with the naive deep ensemble method and show the results in Appendix~\ref{sec:deep ensemble}.

Examining the results, we note that Sup-tickets improve both SST and DST in all settings with a small operation modification of those algorithms. In all settings, a large array of other techniques are outperformed.

\section{Extensive Analysis}

\textbf{Cyclical Length.}  Here, we study how the cyclical length $C$ affects the Sup-tickets' performances. For all experiments, we still take the last 10\% of the training time for the generation of the cheap tickets, while altering the cyclical length as 2, 4, 8, and 12 epochs. The cheap ticket count then varies accordingly. The results are shown in Table~\ref{table:cyc_lenth}. In general, the intermediate lengths (i.e., $C=4$ or $C=8$) tend to achieve better accuracy than the extreme small or large lengths (i.e., $C=2$ or $C=12$). The results are expected since small lengths can not guarantee the high quality (high accuracy) of each cheap ticket, whereas large lengths naturally decrease the number of the collected tickets. Consequently, we use $C=8$ as the default setting in the main experiment section~\ref{sec:experiments sup-tickets}.

\begin{table}[htbp]
\centering
\caption{Test accuracy (\%) on CIFAR-100 of Sup-tickets combined with RigL under different cyclical lengths. The best results are marked in bold. }
\vspace{-0.5em}
\label{table:cyc_lenth}
\resizebox{0.4\textwidth}{!}{
\begin{tabular}{lccc}
\cmidrule[\heavyrulewidth](lr){1-4}

{\textbf{Cyclical}}  & \multicolumn{3}{c}{Pruning ratio } \\   
\cmidrule(lr){2-4}

{\textbf{length (epochs)}} & 95\%      & 90\%     & 80\%         
 \\ 
\cmidrule[\heavyrulewidth](lr){1-4}

\multicolumn{4}{c}{VGG-16} 
\\
\cmidrule[\heavyrulewidth](lr){1-4}
C=2
& 71.35$\pm$0.14 & 72.89$\pm$0.41 & \textbf{73.65$\pm$0.20}
\\
C=4
&\textbf{71.42$\pm$0.19} &\textbf{73.00$\pm$0.20} & 73.62$\pm$0.40
\\
C=8
& 71.31$\pm$0.21 & 72.57$\pm$0.29 &73.61$\pm$0.11
\\
C=12
& 71.27$\pm$0.06 & 72.69$\pm$0.43 &73.45$\pm$0.06
\\
\cmidrule[\heavyrulewidth](lr){1-4}
\multicolumn{4}{c}{ResNet-50} 
\\
\cmidrule[\heavyrulewidth](lr){1-4}
C=2
&\textbf{77.58$\pm$0.22} & 78.48$\pm$0.45 & 78.50$\pm$0.32
\\
C=4
&77.33$\pm$0.26 & \textbf{78.52$ \pm$0.36} & 78.62$\pm$0.34
\\
C=8
&\textbf{77.58$\pm$0.47} & \textbf{78.52$\pm$0.39} & \textbf{78.69$\pm$0.30}
\\
C=12
& 77.17$\pm$0.42 &  78.39$\pm$0.43 & 78.48$\pm$0.38
\\
\cmidrule[\heavyrulewidth](lr){1-4}
\end{tabular}
}
\vspace{-0.5em}
\end{table}

\begin{figure}[h]

\centering

    \subfigure{}{
        \includegraphics[width=0.23\textwidth]{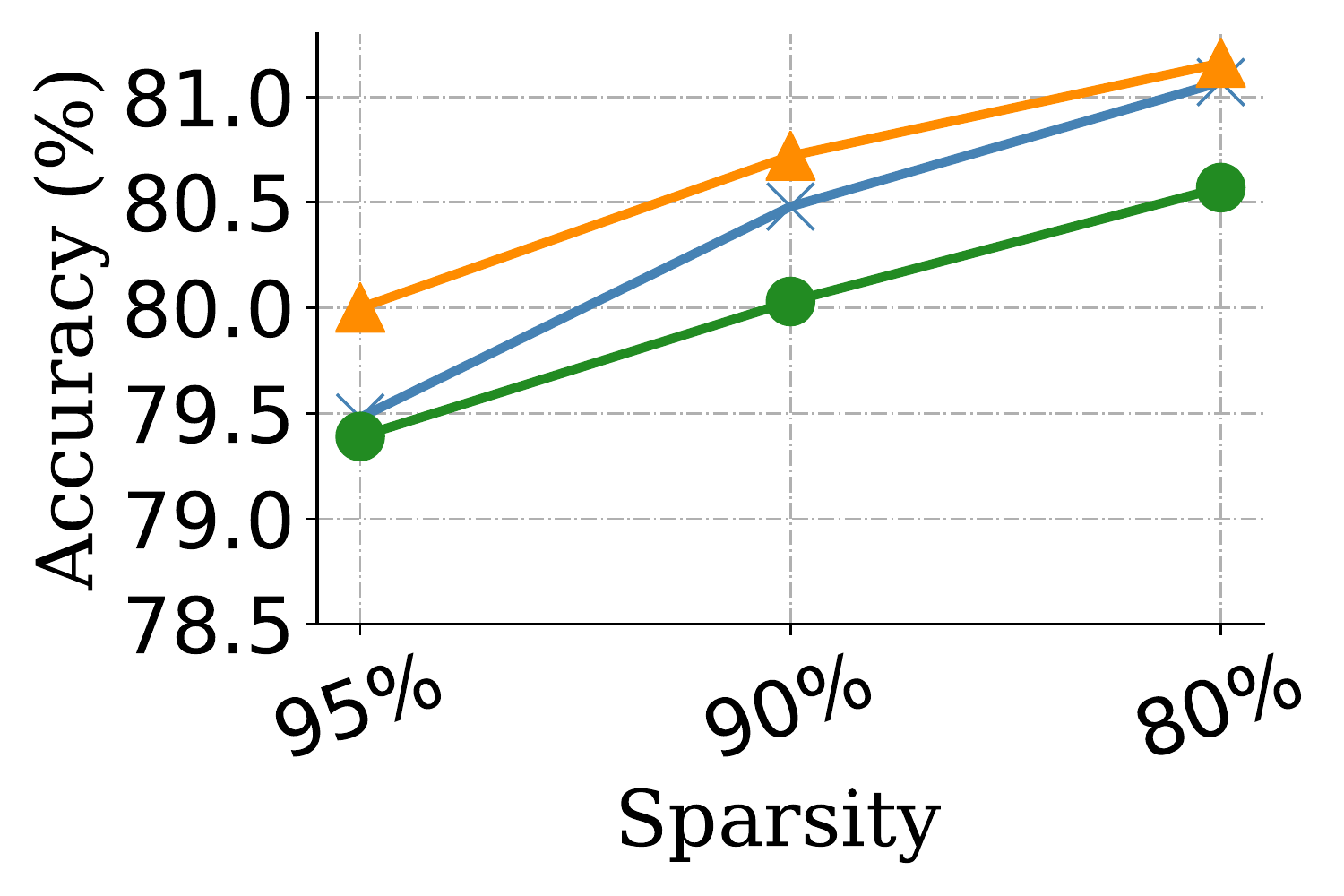}
    }
    \subfigure{}{
        \includegraphics[width=0.23\textwidth]{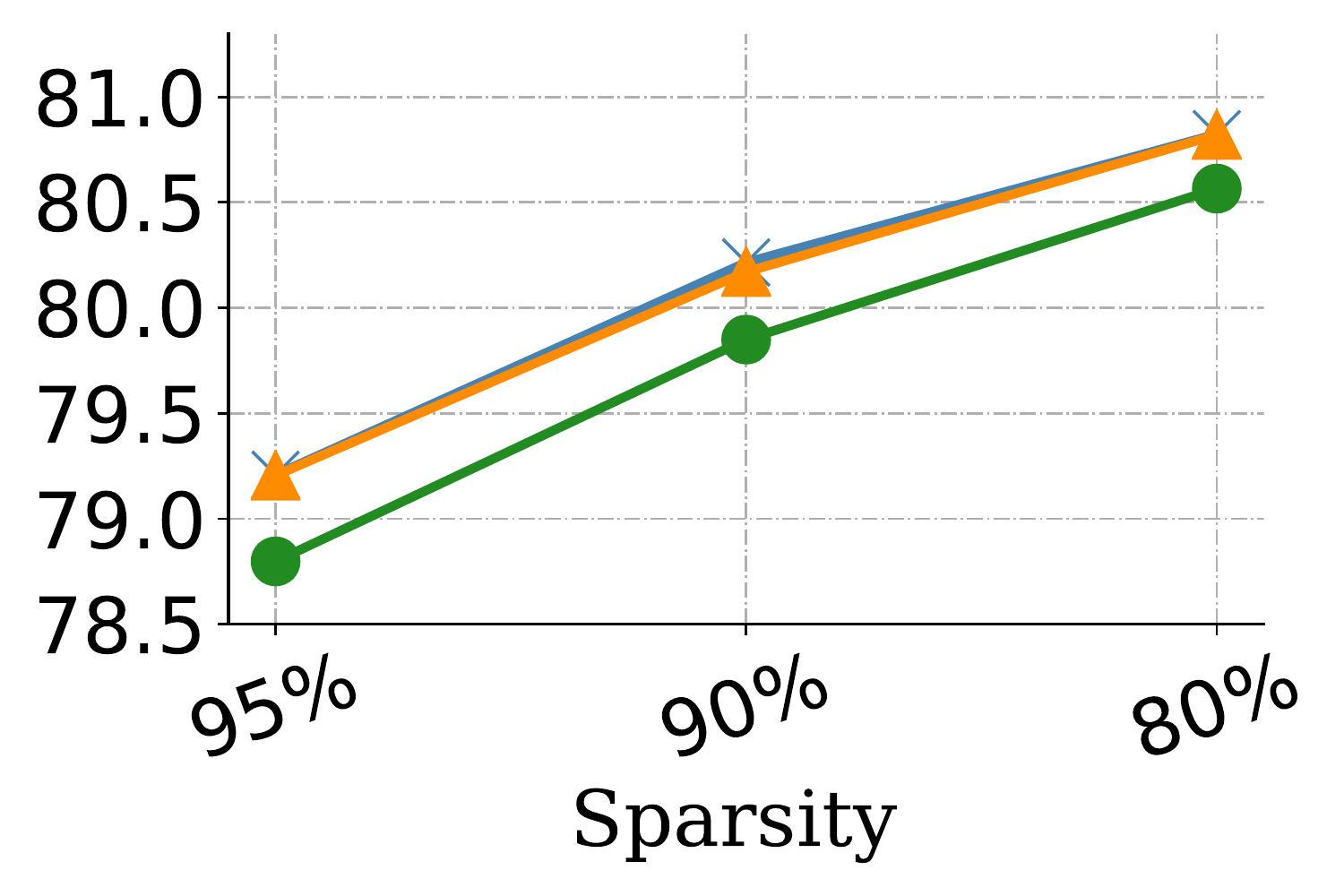}
    }
    \vspace{-0.5em}
    \subfigure{}{
        \includegraphics[width=0.5\textwidth]{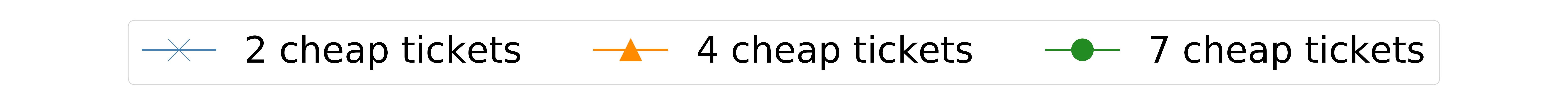}}
\caption{Impacts of the cheap tickets count. Experiments are conducted with Wide ResNet28-10 trained with RigL+Sup-tickets on CIFAR-100. \textbf{Left:} test accuracy of the ultimate tickets.  \textbf{Reft:} the mean accuracy of the individual cheap tickets used to build the ultimate tickets. }

\label{fig:different_tickets_number}

\end{figure}

\looseness=-1 \textbf{Number of Cheap Tickets.} To study the effect of the cheap ticket count on ultimate ticket's performance, we alter the cheap ticket count with 2, 4, and 7, and fix the cyclical length as 8 epochs. The overall training time is set as 250 epochs. Under this setting, the time used for ticket generation is not fixed as 10\%, but it changes according to the cheap ticket count. We report the results in Figure~\ref{fig:different_tickets_number}-left. It could be seen that our approach achieves the best performance under four tickets, not the largest nor the smallest ticket count, apparently since creating too many cheap tickets will reduce the time of the normal sparse training phase, and thus yielding cheap tickets with poor performance. We further prove this in Figure~\ref{fig:different_tickets_number}-right. On the other hand, 2 cheap tickets are too few to boost the performance. Figure~\ref{fig:different_tickets_number} also illustrates the effectiveness of Sup-tickets, where the superposed subnetworks outperform the individual subnetworks by a large margin.

\looseness=-1 {The fixed training time constraint is important to enable comparisons among various sparse training methods since training efficiency is one of the main contributions of sparse training. It is natural to evaluate whether Sup-tickets can lead to continuous improvement when we remove this constraint. To evaluate this, we simply extend the overall training time to yield more cheap tickets. The results are reported in Appendix~\ref{sec:ticket_count}. We can see that the performance of Sup-tickets continuously improves as the number of tickets increases.}

{\textbf{Diversity Analysis.}  We report the diversity of the different subnetworks we obtained during training using KL divergence and prediction disagreement, which are widely used for deep ensembling~\cite{liu2021deep,fort2019deep}. We compare our methods against the traditional dense ensemble and two state-of-the-art efficient ensemble methods, including TreeNet~\citep{lee2015m} and BatchEnsemble~\citep{wen2020batchensemble}, with Wide ResNet28-10 on CIFAR-10. The results are also in line with our intuition. We  observe that the diversity of cheap tickets obtained by our method is lower than the traditional dense ensemble. This makes sense since networks of the traditional dense ensemble are obtained by different runs and should converge to different basins, whereas cheap tickets obtained by our methods are intended to be located in the same basin with relatively lower diversity. Nevertheless, our method still maintains a similar or even higher diversity than TreeNet and BatchEnsemble, verifying its effectiveness. The relatively low diversity ensures that our cheap tickets are located in the same wide and flat low loss region, which is actually crucial for the success of weight averaging, since too diverse networks could lead to very poor performance from the previous experiments~\cite{izmailov2018averaging,wortsman2021learning}. }

\begin{table}[!ht]
\centering
\tiny
\caption{{Prediction disagreement and KL divergence among various  ensemble methods.}}
\vspace{-3mm}
\label{tab:diversity_cheaptickets}

\resizebox{0.45\textwidth}{!}{
\begin{tabular}{lcc}
\toprule
Methods &  {$d_{\text{dis}}$ ($\uparrow$)}  & {$d_{\mathrm{KL}}$  ($\uparrow$)} \\
\midrule
TreeNet~\cite{lee2015m}                    & 0.010 & 0.010 \\ 
BatchEnsemble~\cite{wen2020batchensemble}              & 0.014 & 0.020 \\ 
\cmidrule(lr){1-3}
SET+Sup-tickets  (ours)          & 0.015 & 0.015 \\
Rigl+Sup-tickets  (ours)           & 0.017 & 0.015 \\
\cmidrule(lr){1-3}
Traditional Dense Ensemble             & 0.032 & 0.086 \\

\bottomrule
\end{tabular} }
\end{table}

\begin{figure*}[htbp]
\centering
    \subfigure{}{
        \includegraphics[width=0.23\textwidth]{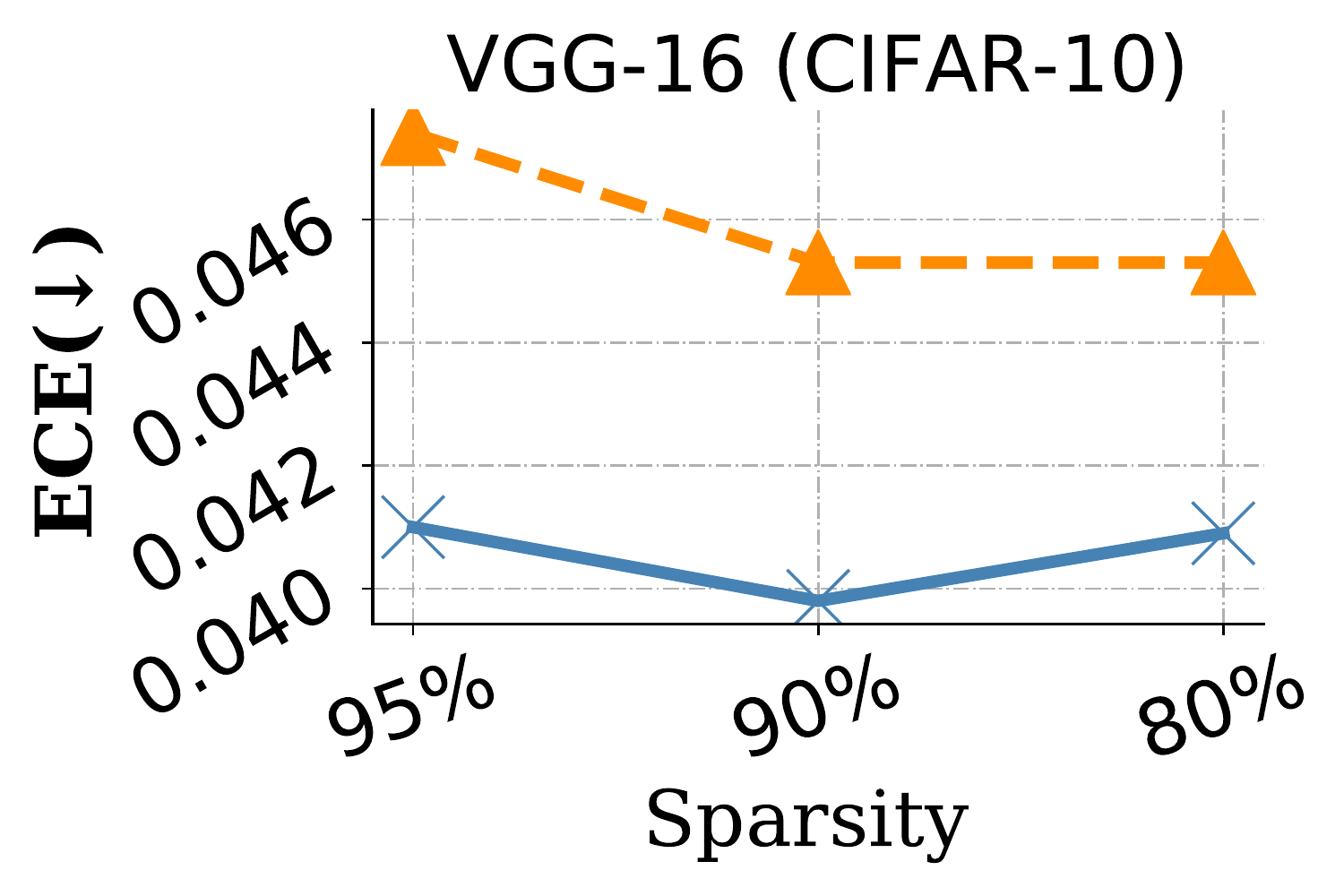}}
    \subfigure{}{
        \includegraphics[width=0.23\textwidth]{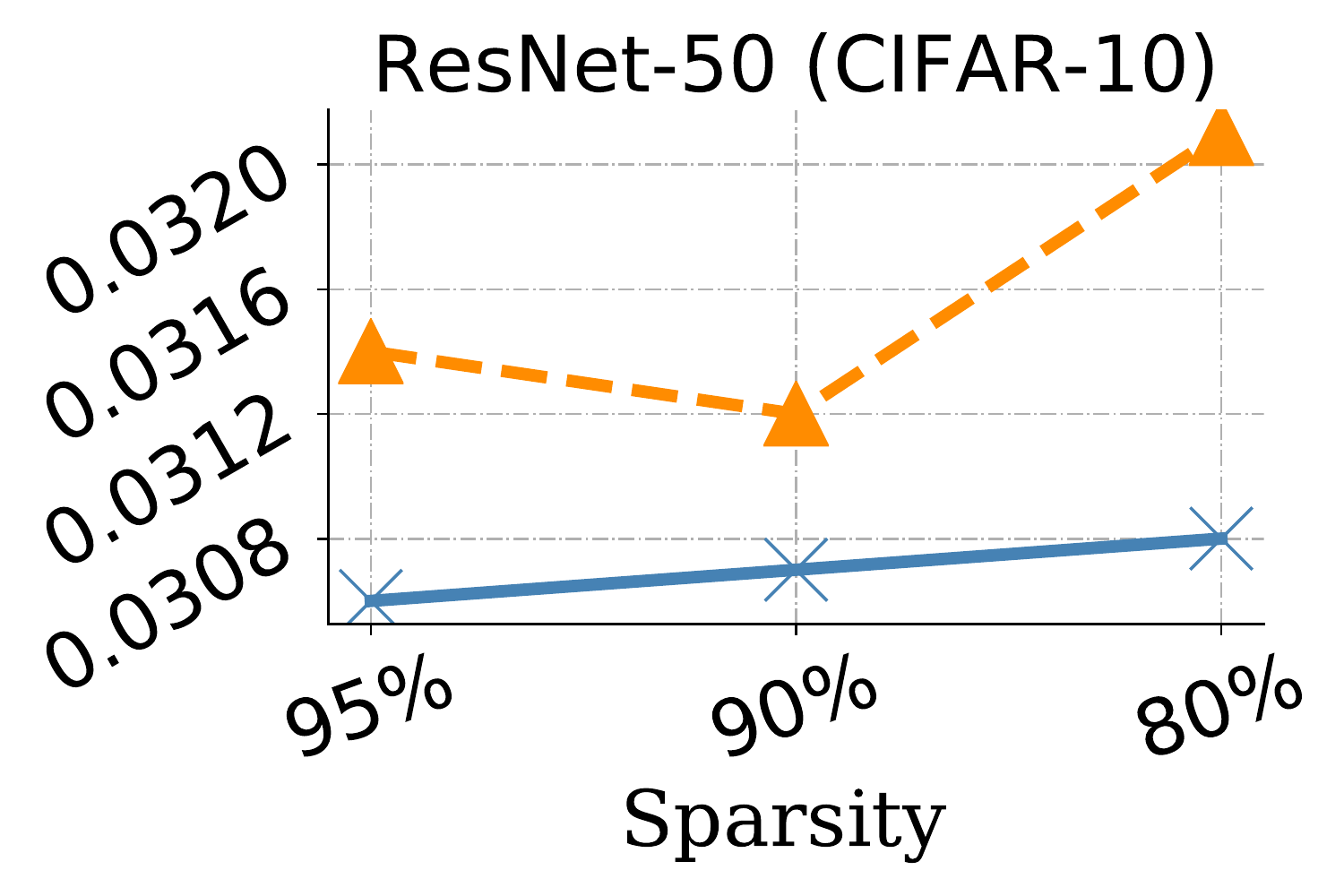}}
    \subfigure{}{
        \includegraphics[width=0.23\textwidth]{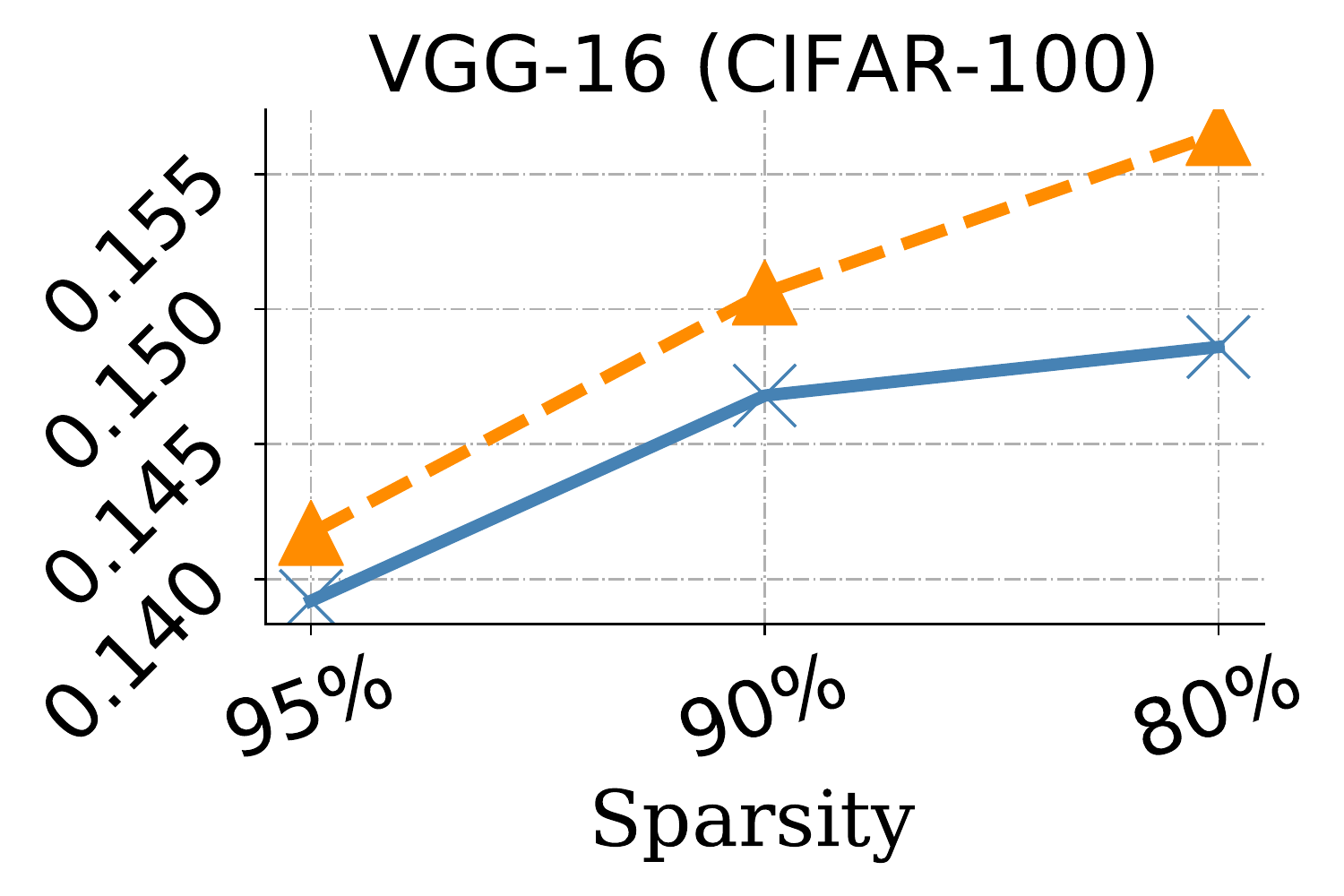}}
    \subfigure{}{
        \includegraphics[width=0.23\textwidth]{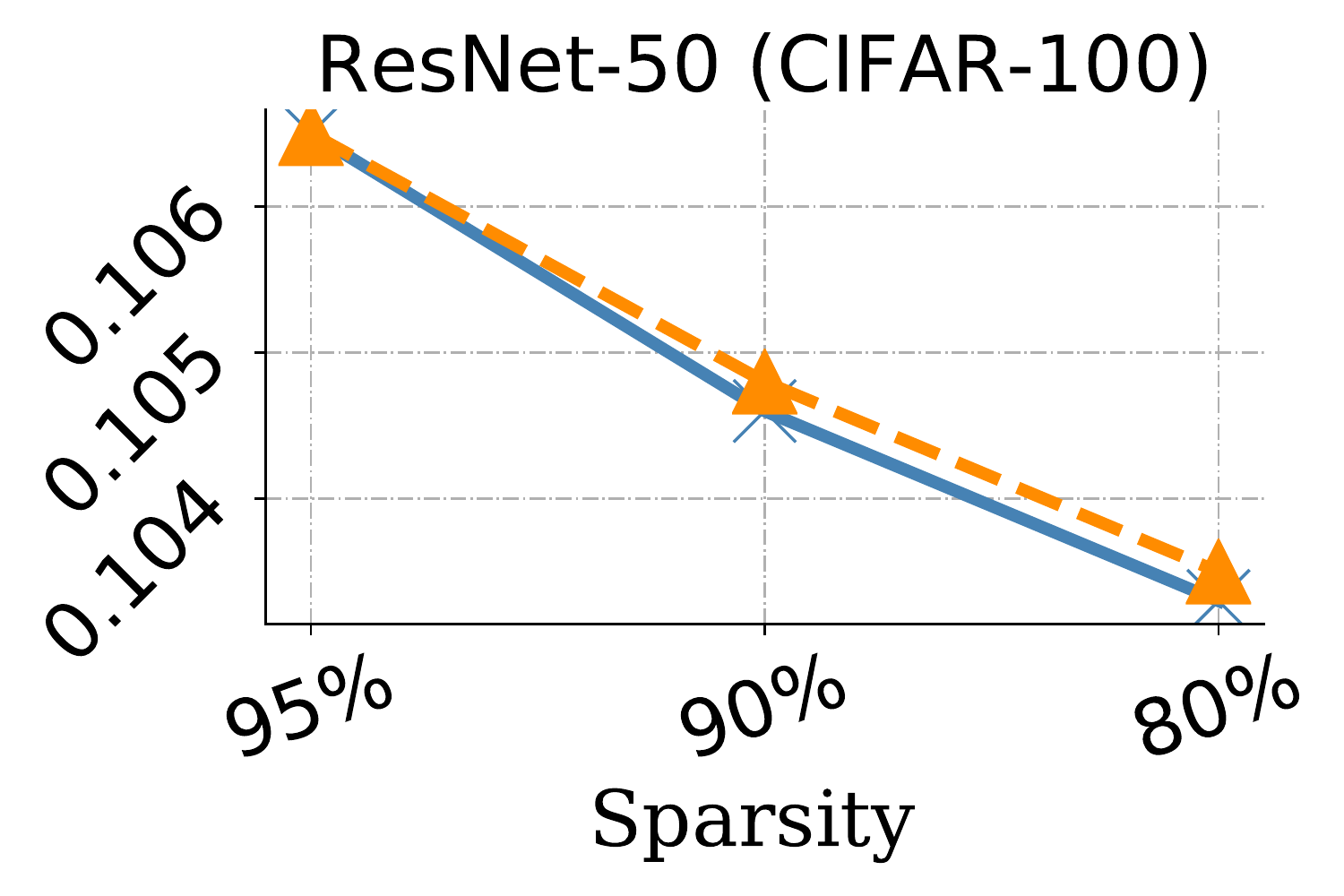}}

    \subfigure{}{
        \includegraphics[width=0.23\textwidth]{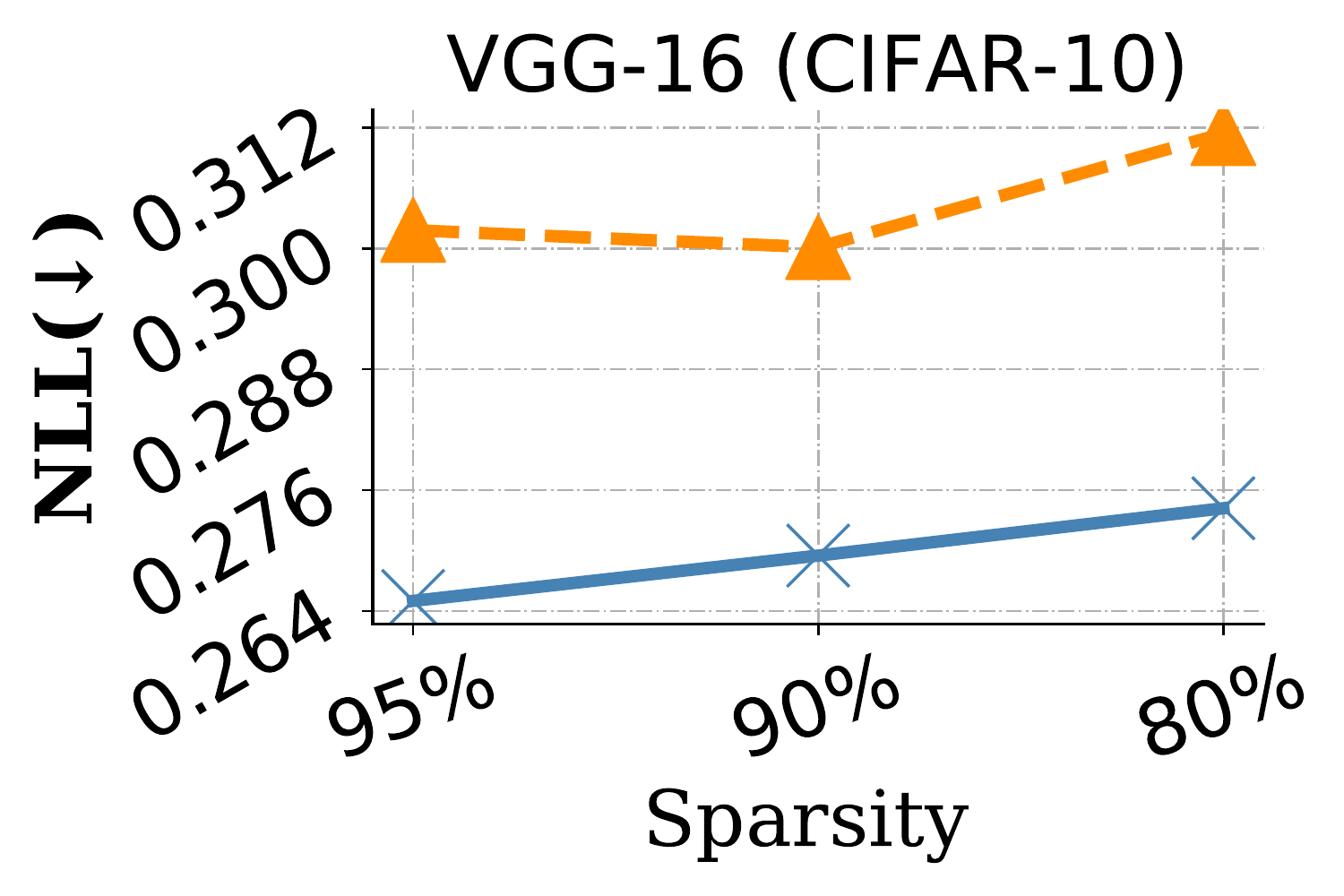}}
    \subfigure{}{
        \includegraphics[width=0.23\textwidth]{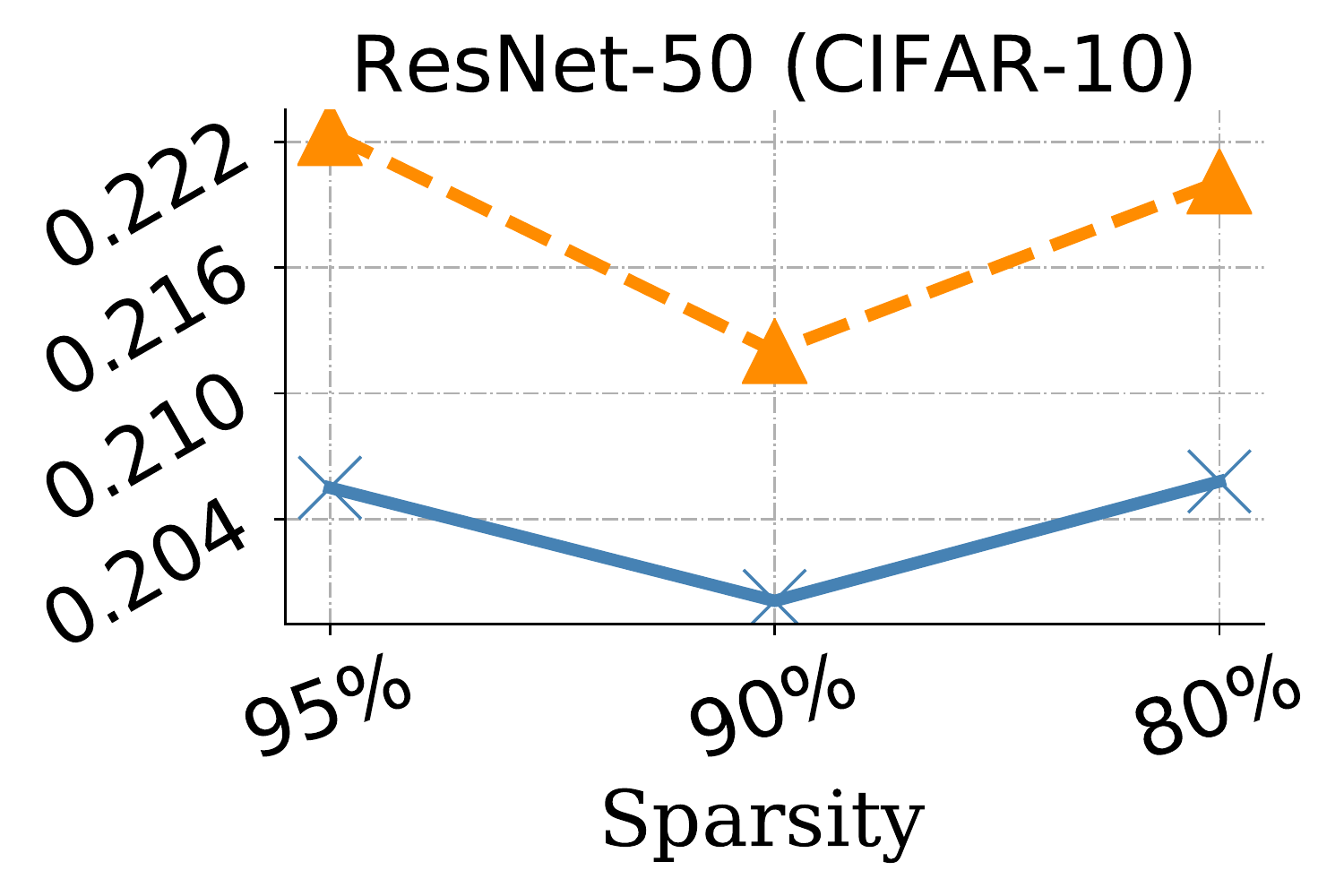}}
    \subfigure{}{
        \includegraphics[width=0.23\textwidth]{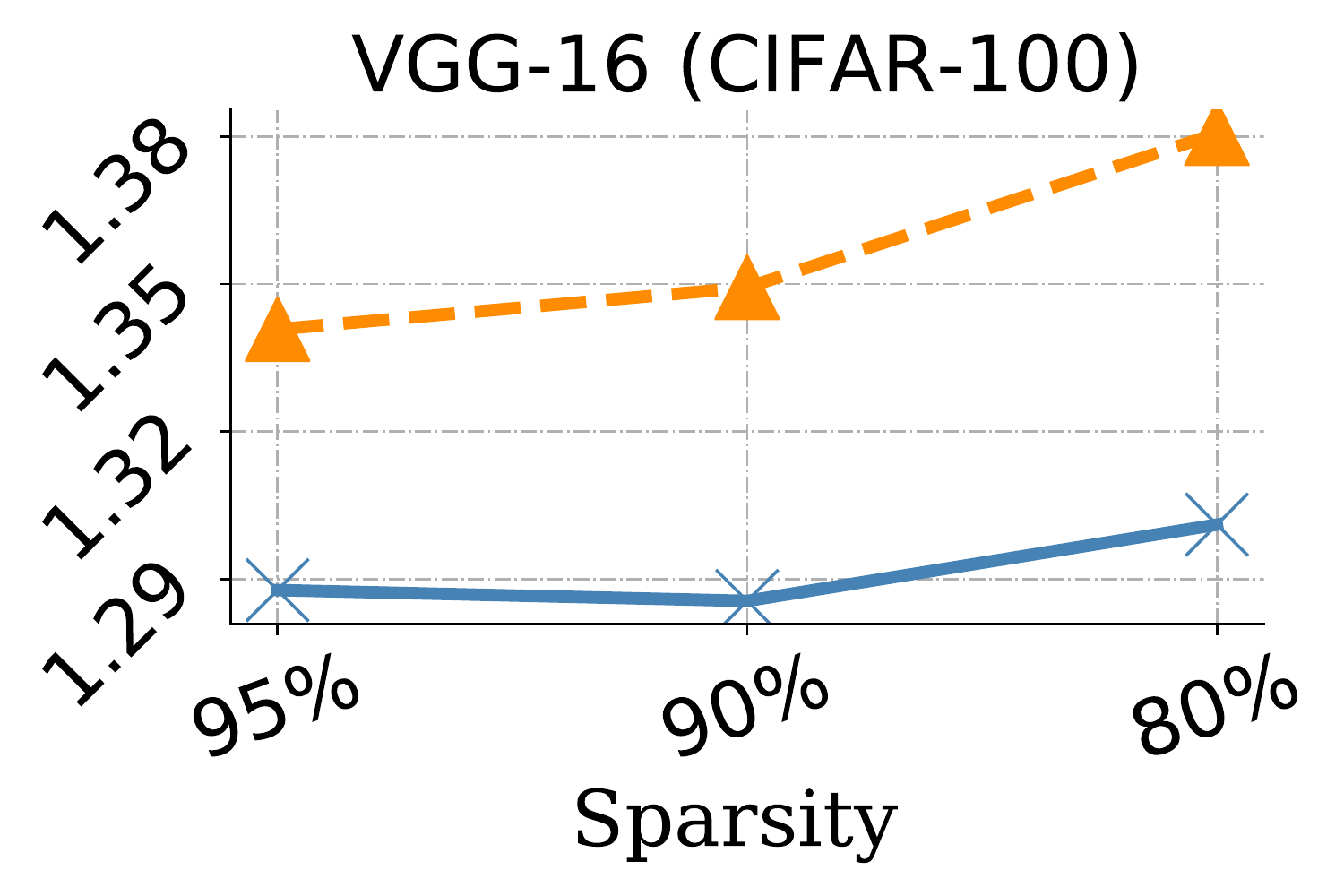}}
    \subfigure{}{
        \includegraphics[width=0.23\textwidth]{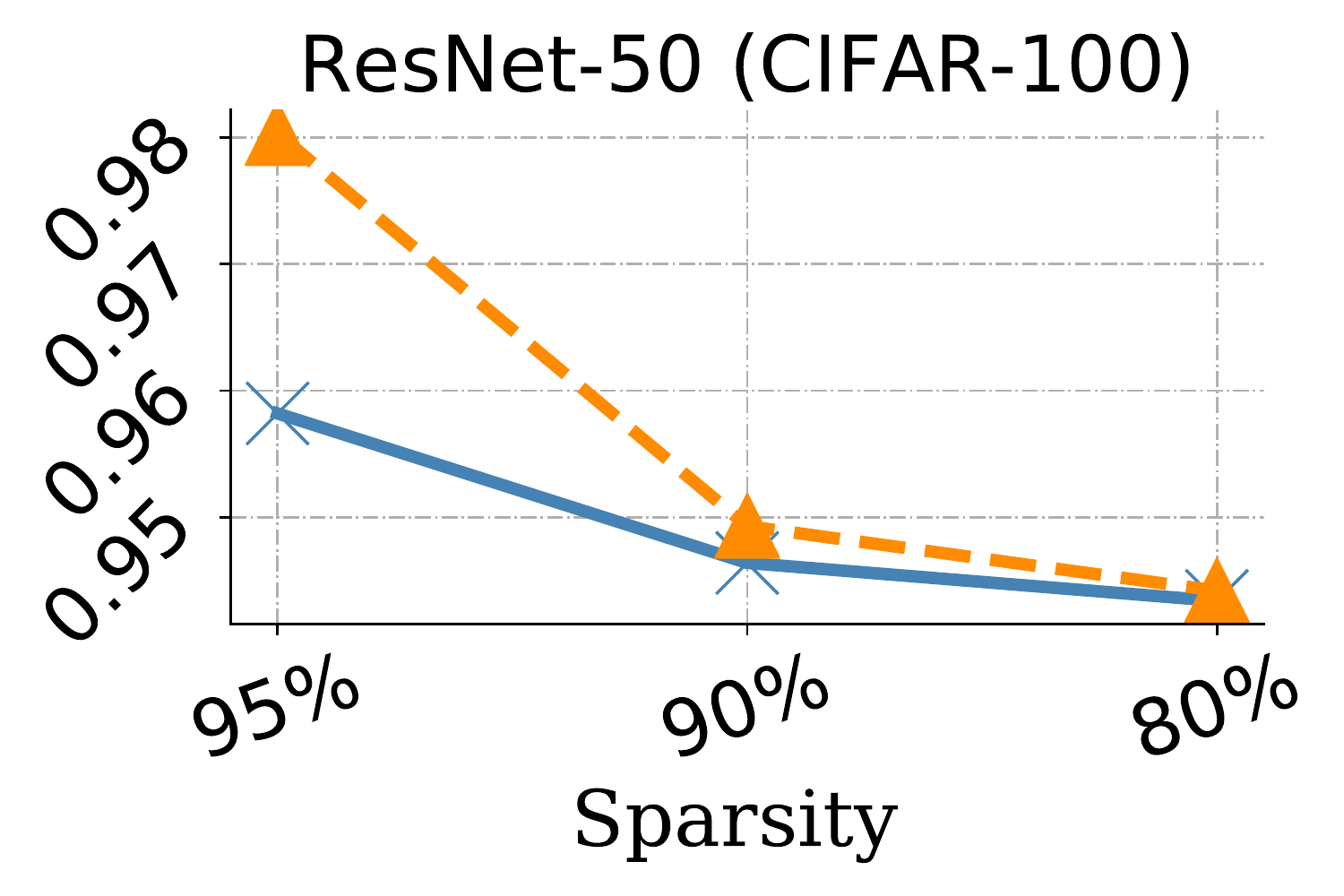}}
    \vspace{-0.5em}
     \subfigure{}{
        \includegraphics[width=0.45\textwidth]{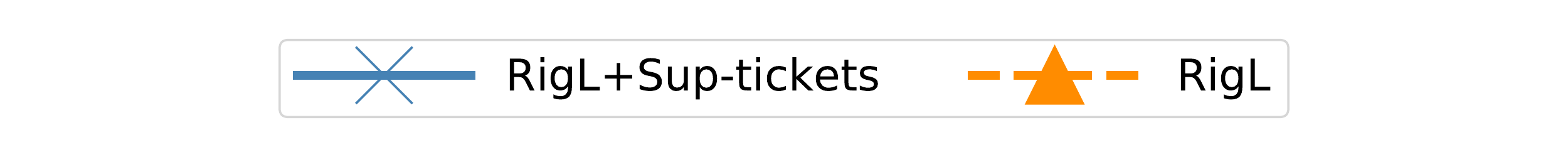}}       
    \vspace{-1.3em}
\caption{Comparison between RigL and RigL+Sup-tickets in terms of ECE and NLL.}
    \label{fig:uncertainty}
\vspace{-1em}
\end{figure*}

{\textbf{Comparison with Different Learning Rate Schedules.}
We compare our method with two learning rate schedule baselines: the learning rate schedule used in FGE~\citep{garipov2018loss} and the learning rate schedule used in SWA~\citep{izmailov2018averaging}. In all learning rate schedules, Sup-tickets are collected at the lowest learning rate stage, and we fixed the learning rate range of these schedules for a fair comparison. Below we report the results on CIFAR-100. All the results are averaged from 3 random runs. It could be seen that our method surpasses the other baselines in 5 out of 6 cases. }

\begin{table}[htbp]
\centering
\caption{{Effect of Various Different Learning Rate (LR) Schedules.}}

\label{table:lr_s}
\resizebox{0.4\textwidth}{!}{
\begin{tabular}{lccc}
\cmidrule[\heavyrulewidth](lr){1-4}

{\textbf{LR schedule}}  & \multicolumn{3}{c}{Sparsity } \\   
\cmidrule(lr){2-4}

{\textbf{Method}} & 95\%      & 90\%     & 80\%         
 \\ 
\cmidrule[\heavyrulewidth](lr){1-4}

\multicolumn{4}{c}{VGG-16} 
\\
\cmidrule[\heavyrulewidth](lr){1-4}
LR of FGE~\citep{garipov2018loss}
& 70.66$ \pm$0.25  & 72.47$\pm$0.44 & 73.22$\pm$0.23
\\
LR of SWA~\citep{izmailov2018averaging}
& 71.26$\pm$0.16& \textbf{72.77$\pm$0.37} & {73.44$\pm$0.19}
\\
Sup-ticket (Ours)
&\textbf{71.31$\pm$0.21} & {72.57$\pm$0.29} &  {\textbf{73.61$\pm$0.11}}
\\
\cmidrule[\heavyrulewidth](lr){1-4}
\multicolumn{4}{c}{ResNet-50} 
\\
\cmidrule[\heavyrulewidth](lr){1-4}
LR of FGE~\citep{garipov2018loss}
&  77.30$\pm$0.67  & 78.20$\pm$0.53 &  78.35$\pm$0.35
\\
LR of SWA~\citep{izmailov2018averaging}
& 77.30$\pm$0.36 & 78.39$\pm$0.38 & 78.48$\pm$0.35
\\
Sup-ticket (Ours)
&\textbf{77.58$\pm$0.47}  &  {\textbf{78.52$\pm$0.39}}  &    {\textbf{78.69$\pm$0.30}}
\\
\cmidrule[\heavyrulewidth](lr){1-4}
\end{tabular}
}
\vspace{-1em}
\end{table}

{We adjust the learning rate schedule slightly so that the learning rate gradually rises to an increased but still small value (0.005) and then decays to the lowest value (0.001) in each cycle. Such a smooth schedule ensures that the new cheap tickets only bounce within the same basin instead of jumping out of it. To help us clarify this, we added extra experiments in which the learning rate will immediately increase to a very large value of 0.1 at the beginning of each cycle. We expect that the large learning rate will force the cheap tickets to jump out of the current basin, and the weight averaging does not bring any performance gains. The results in Table~\ref{table:large_lr} are perfectly in line with our expectations. Parameter averaging significantly degrades the accuracy to 10\% $\sim$ 30\%, even though the accuracy of each subnetwork is still high. Besides, if we generate tickets with different prune/grow criteria, they are also likely located in the different basins and dramatically hurt the performance.}

\begin{table}[!ht]
\centering
\caption{{{Results of Sup-ticket under large restarting learning rate (0.1) and small restarting learning rate (0.005).}}}

\label{table:large_lr}
\resizebox{0.5\textwidth}{!}{
\begin{tabular}{lcccc}
\toprule

Model & Setting &  {Accuracy of Each Ticket}     & Averaged Accuracy \\
\midrule

ResNet-50  & Large LR schedule                    &  [76.05, 76.37, 76.97]   &  10.37 \\ 
          & Low LR schedule (ours)              & [77.15, 77.56, 77.07] & 77.87\\ 
\midrule
VGG-16  & Large LR schedule                      &[68.68, 69.29, 70.36] &  31.74 \\ 
          & Low LR schedule (ours)               & [70.31 ,70.15, 70.32] & 71.19\\
\bottomrule
\end{tabular} }

\end{table}

\textbf{Batch Normalization.} When there are batch normalization (BN) layers~\citep{ioffe2015batch} in the model, traditional weight averaging approaches~\citep{garipov2018loss,izmailov2018averaging} usually run one additional pass over the data to calculate the mean and standard deviation of these layers. Differently, we retrieve these statistics by simply averaging the mean and standard deviation of the BN layers in all cheap tickets without extra forward pass. To avoid extra memory occupation during implementation, similar to the weights averaging operation in Eq.~\ref{eq:average}, we calculate the superposed ticket's BN statistics $\widetilde{\bm{\mathrm{\theta}}}_\mathrm{bn}^\mathrm{t}$  across the first $t$ cheap tickets   using $\frac{(t-1) \cdot \widetilde{\bm{\mathrm{\theta}}}_\mathrm{bn}^\mathrm{t-1} + \bm{\mathrm{\theta}}_\mathrm{bn}^\mathrm{t}}{t}$, where $\bm{\mathrm{\theta}}_\mathrm{bn}^\mathrm{t}$ is the mean and standard deviation from $t^{th}$ cheap ticket's BN layers. The comparison between test accuracy under these two strategies is reported in Appendix~\ref{sec:bn}.

\textbf{Uncertainty Estimation.} In the security-critical scenarios, e.g., self-driving, medical treatment, classifiers should not only 
be accurate but also indicate when they are likely to be incorrect~\citep{guo2017calibration}. We further evaluate the performance of our approach on uncertainty estimation. We choose two widely-used metrics, expected calibration error (ECE)~\citep{guo2017calibration} and negative log-likelihood (NLL)~\citep{quinonero2005evaluating} to enable uncertainty comparisons among different methods. We apply Sup-tickets to RigL and compare it with the vanilla RigL in Figure~\ref{fig:uncertainty}. As observed, in addition to the improvement of accuracy, Sup-tickets also achieves stronger uncertainty estimation performance over RigL, and such improvement can likely generalize to other sparse training methods.

\vspace{-1em}
\section{Conclusion}

In this paper, we presented a novel sparse training approach, Sup-tickets, which effectively produces many cheap subnetworks (tickets) during training and superposes them into one stronger ultimate subnetwork. Sup-tickets is easily combined with existing techniques, agnostic to model architectures, datasets, and is able to boost the sparse training performance with only a negligible amount of extra FLOPs. Across various scenarios, consistent performance improvement is obtained by Sup-tickets in terms of accuracy as well as uncertainty estimation, under the same training time used by the standard sparse training methods. 
It is impressive to see that sup-tickets outperforms the corresponding dense networks on CIFAR-10/100 even in extremely sparse situations when collaborating with GraNet.

There are many potential directions to be explored in the future. For example, even if Sup-tickets enable sparse neural networks to match or outperform their dense counterparts in terms of test accuracy, do they learn the same representation as the latter learn? Besides, we hope the superior performance achieved by Sup-tickets could inspire more researchers to invest in developing hardware accelerators that have better support for sparse training.

\begin{acknowledgements} 
This work used the Dutch national e-infrastructure with the support of the SURF Cooperative using grant no. EINF-2694, EINF-2943 and EINF-2605.

\end{acknowledgements}






\small
\bibliography{ref}

\newpage
\appendix
\onecolumn
\providecommand{\upGamma}{\Gamma}
\providecommand{\uppi}{\pi}

\section{Experimental Results of Wide ResNet28-10 on CIFAR-10/100}
\label{sec:WRN2810}

\begin{table*}[h]
\centering
\caption{Test accuracy (\%) of sparse Wide ResNet28-10 on CIFAR-10/100. All the results are averaged from three random runs. In each setting, the best results are marked in bold.}
\label{table:WRN_CIFAR}
\resizebox{.9\textwidth}{!}{
\begin{tabular}{lccc ccc}
\cmidrule[\heavyrulewidth](lr){1-7}

 \textbf{Dataset}     & \multicolumn{3}{c}{CIFAR-10} & \multicolumn{3}{c}{CIFAR-100}  \\ 
\cmidrule(lr){1-1}
\cmidrule(lr){2-4}
\cmidrule(lr){5-7}
\textbf{Wide ResNet28-10 }~(Dense) 
& 96.00$\pm$0.13  & - & - 
& 81.09$\pm$0.19  & - & - 
\\
\cmidrule(lr){1-1}
\cmidrule(lr){2-4}
\cmidrule(lr){5-7}
Sparsity     & 95\%      & 90\%     & 80\%     
     &  95\%      & 90\%     & 80\%         \\ 
     
\cmidrule(lr){1-1}
\cmidrule(lr){2-4}
\cmidrule(lr){5-7}

SET~\citep{mocanu2018scalable}
& \textbf{95.63$\pm$0.08}  & 95.85$\pm$0.02   & 95.92$\pm$0.25  
& 79.36$\pm$0.14  & 80.44$\pm$0.18   & 80.60$\pm$0.07 
\\
SET+Sup-tickets (ours)
& 95.53$\pm$0.11  & \textbf{95.91$\pm$0.14}  &\textbf{95.93$\pm$0.10}
& \textbf{79.66$\pm$0.18}  & \textbf{80.65$\pm$0.04}   & \textbf{80.91$\pm$0.20}
\\
\cmidrule(lr){1-1}
\cmidrule(lr){2-4}
\cmidrule(lr){5-7}

RigL~\citep{evci2020rigging}
& 95.70$\pm$0.07 & 95.96$\pm$0.12  &  {96.12$\pm$0.05}
& 79.41$\pm$0.24 & 80.45$\pm$0.45 & 80.92$\pm$0.20
\\
RigL+Sup-tickets (ours)
&\textbf{95.90$\pm$0.11} & \textbf{95.98$\pm$0.06}  &  {\textbf{96.15$\pm$0.08}}
&\textbf{80.00$\pm$0.15}  & \textbf{80.72$\pm$0.22} &  {\textbf{81.16$\pm$0.09}}
\\
\cmidrule(lr){1-1}
\cmidrule(lr){2-4}
\cmidrule(lr){5-7}

GraNet~\citep{liu2021neuroregeneration}  
& 95.95$\pm$0.08 &  {96.02$\pm$0.01} &  {\textbf{96.09$\pm$0.07}}
&  80.43$\pm$0.17 & 80.97$\pm$0.16 &  {81.31$\pm$0.09}
\\
GraNet+Sup-tickets (ours)
& {\textbf{96.03$\pm$0.11}} &  {\textbf{96.13$\pm$0.07}} &  {96.08$\pm$0.04}
&\textbf{80.65$\pm$0.06} &  {\textbf{81.20$\pm$0.09}} &  {\textbf{81.42$\pm$0.18}}
\\
\cmidrule[\heavyrulewidth](lr){1-7}
\end{tabular}}
\end{table*}

\section{Impact of the Cheap Tickets without training time constraint  }
\label{sec:ticket_count}

{We extend the overall training time to yield 9 tickets. All the cheap tickets have been trained for 8 epochs. The results on CIFAR-100 are reported below. All results are averaged from 3 random runs. As shown, the performance of Sup-tickets continuously improves as the number of tickets increases. }

\begin{table}[htbp]
\centering
\caption{{Test accuracy (\%) on CIFAR-100 of Sup-tickets combined with RigL under different cheap ticket count. The best results are marked in bold. }}
\vspace{-0.5em}

\label{table:ticket_count}
\resizebox{0.4\textwidth}{!}{
\begin{tabular}{lccc}
\cmidrule[\heavyrulewidth](lr){1-4}

{\textbf{Ticket}}  & \multicolumn{3}{c}{Sparsity } \\   
\cmidrule(lr){2-4}

{\textbf{count}} & 95\%      & 90\%     & 80\%         
 \\ 
\cmidrule[\heavyrulewidth](lr){1-4}

\multicolumn{4}{c}{VGG-16} 
\\
\cmidrule[\heavyrulewidth](lr){1-4}
N=3
 &  71.47$\pm$0.29 &  72.86$\pm$0.22 &  73.42$\pm$0.21
\\
N=6
 &  71.79$\pm$0.10 &  73.19$\pm$0.23 &  73.69$\pm$0.36
\\
N=9
 &  \textbf{71.92$\pm$0.07} &  \textbf{73.30$\pm$0.26} &  \textbf{74.00$\pm$0.38}
\\
\cmidrule[\heavyrulewidth](lr){1-4}
\multicolumn{4}{c}{ResNet-50} 
\\
\cmidrule[\heavyrulewidth](lr){1-4}
N=3
 &  77.14$\pm$0.57 &  77.84$\pm$0.21 &  78.08$\pm$0.40
\\
N=6
 &  77.53$\pm$0.55 &  78.12$\pm$0.32 &  78.18$\pm$0.49
\\
N=9
 & \textbf{77.57$\pm$0.55} &  \textbf{78.15$\pm$0.20} &  \textbf{78.19$\pm$0.46}
\\
\cmidrule[\heavyrulewidth](lr){1-4}
\end{tabular}
}

\end{table}

\section{the variance of the multiple cheap tickets }

{The variance of the cheap tickets obtained by our method is quite low, as shown in the following table. To ensure good final performance, we expect all the subnetworks to be located in the same low-loss basin with similar performance. On the other hand, high variance means that cheap tickets are located in different basins, and weight averaging will not bring performance gains. To verify this hypothesis, we generate 3 cheap subnetworks under 95\% sparsity on CIFAR-100 with high variance by using different prune/grow criteria: prune with high magnitude and grow with high gradient, prune with low magnitude and grow randomly, prune with low magnitude and grow randomly.}

{We find that averaging subnetworks with high variance significantly hurt the performance, likely due to the fact that they are not from the same loss basin. }

\begin{table}[!ht]
\centering
\caption{{ Accuracy (\%)  of each ticket and the averaged ticket under different variance.  }}
\vspace{-0.5em}

\resizebox{0.7\textwidth}{!}{
\begin{tabular}{lcccc}
\toprule

Model & Setting &  {Accuracy of Each Ticket}  &   Variance   & Averaged Accuracy \\
\midrule

ResNet-50  & High Variance                    & [69.44, 76.52, 61.50]  & 6.13 &  2.04 \\ 
          & Low Variance (ours)              & [77.15, 77.56, 77.07] & 0.21 & 77.87\\ 
\midrule
VGG-16  & High Variance                    &[64.02, 70.01, 58.76] & 4.60 &  2.14 \\ 
          & Low Variance (ours)              & [70.31 ,70.15, 70.32] & 0.08 & 71.19\\
\bottomrule
\end{tabular} }
\end{table}

\newpage
\section{Implementation Details  of Sup-Tickets}
\label{sec:implementation}
In this appendix, we report the implementation details for Sup-tickets, including:  total training epochs (T-epochs), epochs of normal sparse training (N-epochs), epochs of cheap tickets generation (C-epochs), length of per cyclical learning rate schedule (C), learning rate (LR), batch size (BS),  learning rate drop (LR Drop), the lowest learning rate of cyclical learning rate schedule (LR-$\alpha_1$), the largest learning rate of cyclical learning rate schedule (LR-$\alpha_2$), weight decay (WD), produced tickets count (Ticket Count), SGD momentum (Momentum), sparse initialization (Sparse Init), etc.

\subsection{Implementation Details for CIFAR-10/100}
.

\begin{table*}[!ht]
\centering
\caption{Implementation hyperparameters of Sup-tickets on CIFAR-10/100}
\label{tab:hypo_hyper_cifar}
\resizebox{1.0\textwidth}{!}{
\begin{tabular}{cccccccccccccccc}
\toprule
Model & T-epochs & N-epochs & C-epochs &C&  BS & LR  & LR Drop, Epochs & LR-$\alpha_2$ & LR-$\alpha_1$ & Ticket Count & Optimizer & WD  & Momentum & Sparse Init & \\ 
\toprule
VGG-16 &  250 & 226 & 24 & 8 & 128  & 0.1 & 10x, [113, 169]  & 0.001 & 0.005   & 3  & SGD &0.9 &5e-4 & ERK  \\
ResNet-50 &  250 & 226 & 24 & 8 &128  & 0.1 & 10x, [113, 169]  & 0.001 & 0.005   & 3  & SGD &0.9 &5e-4 & ERK  \\
Wide ResNet28-10 &  250 & 226 & 24  & 8 & 128  & 0.1 & 10x, [113, 169]  & 0.001 & 0.005   & 3  & SGD &0.9 &5e-4 & ERK  \\
\bottomrule
\end{tabular}}
\end{table*}

\subsection{Implementation Details for ImageNet}

\begin{table*}[!ht]
\centering
\caption{Implementation hyperparameters of Sup-tickets on ImageNet}
\label{tab:hypo_hyper_imgnet}
\resizebox{1.0\textwidth}{!}{
\begin{tabular}{cccccccccccccccc}
\toprule
Model & T-epochs & N-epochs & C-epochs &C & BS & LR  & LR Drop, Epochs & LR-$\alpha_2$ & LR-$\alpha_1$ & Ticket Count & Optimizer & WD & Momentum & Sparse Init & \\ 
\toprule
ResNet-50 &  100 & 92 & 8 &2 & 64  & 0.1 & 10x, [30, 60, 85]  & 0.0001 & 0.0005   & 4  & SGD &0.9 &1e-4 & ERK  \\
\bottomrule
\end{tabular}}
\end{table*}

\newpage
\section{Comparison between different batch normalization updating strategies.}
\label{sec:bn}
In this section, we compare the test accuracy between two batch normalization updating strategies: (1) using additional running pass over the training data; (2) retrieving the statistic by averaging across each cheap ticket (ours). From Table ~\ref{table:imagenet batch normalization imagenet} and Table ~\ref{table:batch normalization cifar},  we find that there is no obvious difference in test accuracy between these two methods. However, our method could save extra computation resources without the additional running pass.

\begin{table}[htbp]
\centering
\caption{Test accuracy (\%) of different batch normalization updating strategies for ResNet 50 on ImageNet. BU stands for batch normalization updating using additional running pass over the data. AV means averaging across each cheap ticket (ours). In each setting, the best results are marked in bold.}
\label{table:imagenet batch normalization imagenet}
\resizebox{0.35\textwidth}{!}{
\begin{tabular}{lcc}
\cmidrule[\heavyrulewidth](lr){1-3}
\textbf{Dataset}     & \multicolumn{2}{c}{ImageNet}  \\ 
\cmidrule[\heavyrulewidth](lr){1-3}
Sparsity      & 90\%     & 80\%        \\ 
\cmidrule(lr){1-1}
\cmidrule(lr){2-3}
RigL+Sup-tickets (AV)
&74.044 & \textbf{75.966}
\\
RigL+Sup-tickets (BU)
&\textbf{74.083}	  &75.925	
\\
\cmidrule(lr){1-1}
\cmidrule(lr){2-3}
GraNet+Sup-tickets (AV)
&74.554  & \textbf{76.168} 
\\
GraNet+Sup-tickets (BU)
&\textbf{74.560}	  & 76.109
\\
\cmidrule[\heavyrulewidth](lr){1-3} 
\end{tabular} }
\end{table}


\begin{table*}[htbp]
\centering
\caption{Test accuracy (\%) of different batch normalization updating strategies on CIFAR-10/100.  BU stands for batch normalization updating using additional running pass over the data. AV means averaging across each cheap ticket (ours). In each setting, the best results are marked in bold.}
\label{table:batch normalization cifar}
\resizebox{0.8\textwidth}{!}{
\begin{tabular}{lccc ccc}
\cmidrule[\heavyrulewidth](lr){1-7}

 \textbf{Dataset}     & \multicolumn{3}{c}{CIFAR-10} & \multicolumn{3}{c}{CIFAR-100}  \\ 
 \cmidrule(lr){1-7}

Sparsity     & 95\%      & 90\%     & 80\%     
     &  95\%      & 90\%     & 80\%         \\ 
     
 \cmidrule(lr){1-1}
\cmidrule(lr){2-4}
\cmidrule(lr){5-7}

\textbf{VGG-16}~(Dense) 
& 93.91$\pm$0.26   & - & - 
& 73.61$\pm$0.45  & - & - 
\\
SET+Sup-tickets (AV)
&{93.22$\pm$0.09}  & \textbf{93.63$\pm$0.05} & {93.80$\pm$0.13}
&{71.18$\pm$0.29}  & \textbf{71.99$\pm$0.27} & {73.02$\pm$0.32}
\\
SET+Sup-tickets (BU)
&{93.22$\pm$0.12}  & {93.62$\pm$0.01} & {93.80$\pm$0.01}
&\textbf{71.30$\pm$0.26}  & {71.96$\pm$0.19} & \textbf{73.04$\pm$0.31}
\\
\cmidrule(lr){1-1}
\cmidrule(lr){2-4}
\cmidrule(lr){5-7}

RigL+Sup-tickets (AV)
&{93.20$\pm$0.13}  &  {93.81$\pm$0.11}  & {93.85$\pm$0.25}
&{71.31$\pm$0.21} & {72.57$\pm$0.29} & {73.61$\pm$0.11}
\\
RigL+Sup-tickets (BU)
&\textbf{93.24$\pm$0.11}  &  \textbf{93.86$\pm$0.15}  & \textbf{93.88$\pm$0.28}
&\textbf{71.36$\pm$0.16} & \textbf{72.60$\pm$0.27} & \textbf{73.68$\pm$0.16}
\\
\cmidrule(lr){1-1}
\cmidrule(lr){2-4}
\cmidrule(lr){5-7}
GraNet+Sup-tickets (AV)
&{94.10$\pm$0.06} & \textbf{94.13$\pm$0.12} & {94.24$\pm$0.05}
& { 73.61$\pm$0.24}& \textbf{73.87$\pm$0.26} & {73.95$\pm$0.30}
\\
GraNet+Sup-tickets (BU)
&\textbf{94.14$\pm$0.06} & {94.10$\pm$0.14} & \textbf{94.25$\pm$0.07}
& \textbf{ 73.71$\pm$0.21}& {73.79$\pm$0.21} & \textbf{74.03$\pm$0.27}
\\
\cmidrule[\heavyrulewidth](lr){1-7}
\textbf{Wide ResNet28-10 }~(Dense) 
& 96.00$\pm$0.13  & - & - 
& 81.09$\pm$0.19  & - & - 
\\
SET+Sup-tickets (AV)
& 95.53$\pm$0.11  & {95.91$\pm$0.14}  &{95.92$\pm$0.10}
& \textbf{79.66$\pm$0.18}  & \textbf{80.65$\pm$0.04}   & \textbf{80.91$\pm$0.20}
\\
SET+Sup-tickets (BU)
&  \textbf{95.59$\pm$0.11}  & \textbf{95.98$\pm$0.08}  &\textbf{95.97$\pm$0.06}
&79.36$\pm$0.35  & {80.47$\pm$0.05}   & {80.74$\pm$0.21}
\\
\cmidrule(lr){1-1}
\cmidrule(lr){2-4}
\cmidrule(lr){5-7}

RigL+Sup-tickets (AV)
&\textbf{95.90$\pm$0.11} & \textbf{95.98$\pm$0.06}  & {96.15$\pm$0.08}
&\textbf{80.00$\pm$0.15}  & \textbf{80.72$\pm$0.22} & \textbf{81.16$\pm$0.09}
\\
RigL+Sup-tickets (BU)
&{95.88$\pm$0.10} & {95.97$\pm$0.04}  & \textbf{96.17$\pm$0.11}
&{79.76$\pm$0.23}  & {80.52$\pm$0.20} & {81.13$\pm$0.15}
\\
\cmidrule(lr){1-1}
\cmidrule(lr){2-4}
\cmidrule(lr){5-7}

GraNet+Sup-tickets (AV)
&\textbf{96.03$\pm$0.11} & 96.13$\pm$0.07 & 96.08$\pm$0.04
&{80.65$\pm$0.06} & \textbf{81.20$\pm$0.09} & \textbf{81.42$\pm$0.18}
\\

GraNet+Sup-tickets (BU)
&{96.01$\pm$0.07} & \textbf{96.19$\pm$0.08} & \textbf{96.14$\pm$0.09}
&\textbf{80.73$\pm$0.04} & {81.17$\pm$0.13} & {81.39$\pm$0.21}
\\

\cmidrule[\heavyrulewidth](lr){1-7}

\textbf{ ResNet-50 }~(Dense) 
& 94.88$\pm$0.11 & - & - 
& 78.00$\pm$0.40  & - & - 
\\
\cmidrule[\heavyrulewidth](lr){1-7}

SNIP+Sup-tickets (AV)
& {94.33$\pm$0.09} & {95.05$\pm$0.22} & {95.21$\pm$0.09}
& \textbf{65.56$\pm$1.15} & {76.34$\pm$0.27} & \textbf{77.43$\pm$0.53}
\\
SNIP+Sup-tickets (BU)
& \textbf{94.39$\pm$0.06} & \textbf{95.10$\pm$0.12} & \textbf{95.30$\pm$0.02}
& {65.51$\pm$0.83} & \textbf{76.62$\pm$0.23} & {77.35$\pm$0.62}
\\
\cmidrule(lr){1-1}
\cmidrule(lr){2-4}
\cmidrule(lr){5-7}
ERK+Sup-tickets (AV)
& {93.92$\pm$0.04} & {94.80$\pm$0.06} & {95.11$\pm$0.27}
& {75.75$\pm$0.28} & {76.82$\pm$0.08} & \textbf{77.85$\pm$0.42}
\\
ERK+Sup-tickets (BU)
& \textbf{93.99$\pm$0.08} & \textbf{94.87$\pm$0.04} & \textbf{95.18$\pm$0.27}
& \textbf{76.02$\pm$0.22} & \textbf{77.01$\pm$0.17} & {77.80$\pm$0.54}
\\
\cmidrule(lr){1-1}
\cmidrule(lr){2-4}
\cmidrule(lr){5-7}

SET+Sup-tickets (AV)
&{94.81$\pm$0.05}   & {94.87$\pm$0.03}  & \textbf{94.90$\pm$0.27}
&\textbf{ 76.68$\pm$0.38}  & {77.89$\pm$0.45}   & 78.35$\pm$0.18
\\
SET+Sup-tickets (BU)
&\textbf{94.85$\pm$0.03}   & \textbf{94.97$\pm$0.05}  & {94.86$\pm$0.20}
&{ 76.54$\pm$0.41}  & \textbf{77.93$\pm$0.50}   & \textbf{78.38$\pm$0.18}
\\
\cmidrule(lr){1-1}
\cmidrule(lr){2-4}
\cmidrule(lr){5-7}

RigL+Sup-tickets (AV)
& \textbf{94.65$\pm$0.11}  & {94.82$\pm$0.13} &  \textbf{94.81$\pm$0.15}
&\textbf{77.58$\pm$0.47}  &  \textbf{78.52$\pm$0.39}  &   \textbf{78.69$\pm$0.30}
\\
RigL+Sup-tickets (BU)
&{94.64$\pm$0.13}  &  \textbf{94.89$\pm$0.09}  &   {94.79$\pm$0.17}
&{77.54$\pm$0.53}  &  {78.43$\pm$0.40}  &   {78.53$\pm$0.31}
\\

\cmidrule(lr){1-1}
\cmidrule(lr){2-4}
\cmidrule(lr){5-7}

GraNet+Sup-tickets (AV)
&{94.89$\pm$0.15} & {95.08$\pm$0.08} & {94.94$\pm$0.03}
&{77.70$\pm$0.47} & {78.37$\pm$0.53}  & \textbf{78.95$\pm$0.33}

\\
GraNet+Sup-tickets (BU)
&\textbf{94.91$\pm$0.19} & \textbf{95.16$\pm$0.14} & \textbf{95.09$\pm$0.03}
&\textbf{77.82$\pm$0.60} & \textbf{78.63$\pm$0.64}  & {78.07$\pm$0.32}

\\

\cmidrule[\heavyrulewidth](lr){1-7}
\end{tabular}
}

\end{table*}

 \newpage
\section{Layer-wise Sparsity of ResNet-50 on ImageNet}

Table~\ref{tab:res50sparsity} summarizes the final sparsity budgets for 90\% sparse ResNet-50 on ImageNet-1K obtained by various methods. Backbone represents the sparsity budgets for all the CNN layers without the last fully-connected layer.

\begin{table}[!ht]
\centering
\caption{ResNet-50 Learnt Budgets and Backbone Sparsities at Sparsity 90\%
}
\label{tab:res50sparsity}
\resizebox{0.8\columnwidth}{!}{
\begin{tabular}{@{}l|rr|cccccccc@{}}
\toprule
\multirow{2}{*}{Metric}          & \multicolumn{1}{c}{\multirow{2}{*}{\begin{tabular}[c]{@{}c@{}}Fully Dense \\ Params\end{tabular}}} & \multicolumn{1}{c|}{\multirow{2}{*}{\begin{tabular}[c]{@{}c@{}}Fully Dense \\ FLOPs\end{tabular}}} & \multicolumn{7}{c}{Sparsity (\%)}                     \\ \cmidrule(l){4-11} 
                                 & \multicolumn{1}{c}{}                                                                               & \multicolumn{1}{c|}{}                                                                              &  GraNet+Sup-tickets & GraNet  & RigL+Sup-tickets &RigL   \\ \midrule
Overall & 25502912 & 8178569216 & 89.99 & 89.98 & 90.23   & 90.00  \\
Backbone & 23454912 & 8174272512 & 89.89 & 90.65 & 92.47 
& 90.00 
\\ \midrule
Layer 1 - conv1 & 9408 & 118013952 & 37.40 & 38.22 & 57.26 & 58.32
\\
Layer 2 - layer1.0.conv1 & 4096 & 236027904 & 40.55 & 41.70 & 14.58 & 9.40
\\
Layer 3 - layer1.0.conv2 & 36864 & 231211008 & 64.88 & 65.05 & 82.13 & 82.40
\\
Layer 4 - layer1.0.conv3 & 16384 & 102760448 & 64.69 & 65.09 & 17.13 & 16.41
\\
Layer 5 - layer1.0.downsample.0 & 16384 & 102760448 & 74.75 & 74.99 & 29.10 & 24.25
\\
Layer 6 - layer1.1.conv1 & 16384 & 102760448 & 66.33 & 66.75 & 19.72 & 19.02
\\
Layer 7 - layer1.1.conv2 & 36864 & 231211008 & 62.25 & 62.62 & 82.05 & 82.44
\\
Layer 8 - layer1.1.conv3 & 16384 & 102760448 & 57.99 & 58.57 & 4.79 & 4.07
\\
Layer 9 - layer1.2.conv1 & 16384  & 102760448  & 60.15 & 60.60 & 4.85 & 4.19
\\
Layer 10 - layer1.2.conv2 & 36864 & 231211008 & 57.15 & 57.45 & 81.73 & 82.06
\\
Layer 11 - layer1.2.conv3 & 16384 & 102760448  & 57.10 & 57.47 & 5.13 & 3.88   
\\
Layer 12 - layer2.0.conv1 & 32768 & 205520896 & 49.90 & 50.42 & 41.61 & 42.37
\\
Layer 13 - layer2.0.conv2 & 147456  & 231211008 & 69.44 & 69.49 & 91.09 & 91.25 
\\
Layer 14 - layer2.0.conv3 & 65536 & 102760448 & 60.42 & 60.74 & 51.43 & 51.98
\\
Layer 15 - layer2.0.downsample.0 & 131072 & 205520896 & 87.23 & 87.26 & 71.36 & 71.27
\\
Layer 16 - layer2.1.conv1 & 65536  & 102760448  & 84.79 & 84.91 & 52.47 & 52.40  
\\
Layer 17 - layer2.1.conv2  & 147456 & 231211008  & 83.03 & 83.07 & 91.25 & 91.34
\\
Layer 18 - layer2.1.conv3  & 65536 & 102760448 & 70.03 & 70.25 & 52.06 & 52.43 
\\
Layer 19 - layer2.2.conv1 & 65536& 102760448 & 79.47 & 79.61 & 52.07 & 52.25
\\
Layer 20 - layer2.2.conv2  & 147456 & 231211008 & 81.78 & 81.82 & 91.28 & 91.38
\\
Layer 21 - layer2.2.conv3  & 65536 & 102760448 & 73.76 & 73.92 & 51.76 & 51.95
\\
Layer 22 - layer2.3.conv1 & 65536 & 102760448 & 74.82 & 74.97 & 51.92 & 52.24 
\\
Layer 23 - layer2.3.conv2 & 147456 & 231211008 & 82.78 & 82.81 & 91.22 & 91.33 
\\
Layer 24 - layer2.3.conv3 & 65536 & 102760448  & 76.61 & 76.73 & 51.86 & 52.01 
\\
Layer 25 - layer3.0.conv1 & 131072 & 205520896 & 60.53 & 60.81 & 70.98 & 71.39
\\
Layer 26 - layer3.0.conv2 & 589824 & 231211008 & 83.45 & 83.41 & 95.66 & 95.72  
\\
Layer 27 - layer3.0.conv3 & 262144 & 102760448 & 69.56 & 69.73 & 75.77 & 76.06
\\
Layer 28 - layer3.0.downsample.0 & 524288 & 205520896 & 95.24 & 95.21 & 85.79 & 85.64
\\
Layer 29 - layer3.1.conv1 & 262144 & 102760448 & 91.19 & 91.22 & 76.02 & 76.03
\\
Layer 30 - layer3.1.conv2 & 589824 & 231211008 & 92.86 & 92.87 & 95.68 & 95.73
\\
Layer 31 - layer3.1.conv3 & 262144 & 102760448  & 80.70 & 80.81 & 75.76 & 75.95  
\\
Layer 32 - layer3.2.conv1 & 262144 & 102760448  & 90.34 & 90.40 & 76.09 & 76.18  
\\
Layer 33 - layer3.2.conv2 & 589824 & 231211008 & 93.22 & 93.24 & 95.68 & 95.73 
\\
Layer 34 - layer3.2.conv3 & 262144 & 102760448 & 83.42 & 83.47 & 76.06 & 76.21  
\\
Layer 35 - layer3.3.conv1 & 262144 & 102760448 & 89.12 & 89.17 & 76.14 & 76.23
\\
Layer 36 - layer3.3.conv2 & 589824 & 231211008 & 93.20 & 93.21 & 95.67 & 95.71
\\
Layer 37 - layer3.3.conv3 & 262144 & 102760448 & 86.26 & 86.30 & 76.13 & 76.24
\\
Layer 38 - layer3.4.conv1 & 262144 & 102760448  & 88.64 & 88.70 & 75.85 & 75.97
\\
Layer 39 - layer3.4.conv2 & 589824 & 231211008 & 94.50 & 94.51 & 95.65 & 95.69 
\\
Layer 40 - layer3.4.conv3 & 262144 & 102760448 & 87.05 & 87.09 & 75.94 & 76.05
\\
Layer 41 - layer3.5.conv1 & 262144 & 102760448 & 87.10 & 87.15 & 75.91 & 76.07
\\
Layer 42 - layer3.5.conv2 & 589824 & 231211008 & 95.13 & 95.14 & 95.69 & 95.72
\\
Layer 43 - layer3.5.conv3 & 262144 & 102760448 & 88.91 & 88.95 & 76.06 & 76.14 
\\
Layer 44 - layer4.0.conv1 & 524288 & 205520896  & 72.04 & 72.13 & 85.54 & 85.67 
\\
Layer 45 - layer4.0.conv2 & 2359296 & 231211008 & 93.56 & 93.53 & 97.84 & 97.86
\\
Layer 46 - layer4.0.conv3 & 1048576 & 51380224 & 82.00 & 82.01 & 88.01 & 88.09 
\\
Layer 47 - layer4.0.downsample.0 & 2097152 & 205520896 & 99.25 & 99.24 & 92.96 & 92.84 
\\
Layer 48 - layer4.1.conv1 & 1048576 & 102760448 & 95.73 & 95.74 & 88.02 & 88.07
\\
Layer 49 - layer4.1.conv2 & 2359296 & 231211008  & 97.39 & 97.39 & 97.86 & 97.87
\\
Layer 50 - layer4.1.conv3 & 1048576 & 102760448 & 91.08 & 91.07 & 88.10 & 88.12
\\
Layer 51 - layer4.2.conv1 & 1048576 & 205520896 & 87.68 & 87.70 & 87.99 & 88.04  
\\
Layer 52 - layer4.2.conv2 & 2359296 & 231211008  & 97.02 & 97.01 & 97.86 & 97.86
\\
Layer 53 - layer4.2.conv3 & 1048576 & 102760448 & 84.54 & 84.50 & 88.07 & 88.07
\\
Layer 54 - fc & 2048000 & 4096000 & 82.70 & 82.54 & 92.78 & 92.74
\\ 
\bottomrule
\end{tabular}}
\end{table}

\newpage
\section{Comparison with outputs ensemble and knowledge distillation}
\label{sec:deep ensemble}
This appendix compares our approach with the prediction ensemble (averaging prediction of subnetworks). For deep ensemble, we use the same procedure to generate cheap tickets as in Sup-tickets; but instead of averaging their weights and connection topology, we save all the cheap tickets in memory and average their softmax outputs at inference stage~\citep{huang2017snapshot,garipov2018loss}. 

The results are reported in Table~\ref{table:test_ensemble_cifar}  $\&$ Table~\ref{table:test_ensemble_imagnetnet}. Across extensive settings, we observe that our sup-tickets could closely match the strong baseline of output averaging. Worth noting that compared with the latter, our method does not require performing multiple forward passes for prediction nor saving all the ensemble members.

\begin{table*}[htbp]
\centering
\caption{{\textbf{Comparison with prediction ensemble.} Test accuracy (\%) of Sup-tickets and naive deep ensemble on CIFAR10/100.}}
\label{table:test_ensemble_cifar}
\resizebox{0.8\textwidth}{!}{
\begin{tabular}{lccc ccc}
\cmidrule[\heavyrulewidth](lr){1-7}

 \textbf{Dataset}     & \multicolumn{3}{c}{CIFAR-10} & \multicolumn{3}{c}{CIFAR-100}  \\ 
 \cmidrule(lr){1-1}
\cmidrule(lr){2-4}
\cmidrule(lr){5-7}

Sparsity     & 95\%      & 90\%     & 80\%     
     &  95\%      & 90\%     & 80\%         \\ 

\cmidrule(lr){1-1}
\cmidrule(lr){2-4}
\cmidrule(lr){5-7}
VGG-16
\\ 
\cmidrule(lr){1-7}
RigL + Prediction Ensemble
& \textbf{93.25$\pm$0.18} &  \textbf{93.82$\pm$0.09} & \textbf{93.97$\pm$0.18}
& \textbf{71.80$\pm$0.24} & \textbf{73.07$\pm$0.34} & \textbf{73.80$\pm$0.21}
\\
RigL + Sup-tickets (ours)
&{93.20$\pm$0.13}  &  {93.81$\pm$0.11}  & {93.85$\pm$0.25}
&{71.31$\pm$0.21} & {72.57$\pm$0.29} &  {{73.61$\pm$0.11}}
\\
\cmidrule(lr){1-7}
ResNet-50
\\ 
\cmidrule(lr){1-7}
RigL + Prediction Ensemble
 &94.64$\pm$0.12 & \textbf{94.94$\pm$0.06} & \textbf{94.86$\pm$0.25 }
 &\textbf{77.66$\pm$0.4} & \textbf{78.54$\pm$0.41} & 78.67$\pm$0.25
\\
RigL + Sup-tickets (ours)
& \textbf{94.65$\pm$0.11}  & {94.82$\pm$0.13} &  {94.81$\pm$0.15}
&{77.58$\pm$0.47}  &  {{78.52$\pm$0.39}}  &    \textbf{78.69$\pm$0.30}
\\
\cmidrule(lr){1-7}
\\

\end{tabular}}
\end{table*}

\begin{table}[htbp]
\centering
\caption{Test accuracy (\%) of Sup-tickets and naive deep ensemble for  ResNet-50 on ImageNet. In each setting, the best results are marked in bold.}
\label{table:test_ensemble_imagnetnet}
\resizebox{0.35\textwidth}{!}{
\begin{tabular}{lcc}
\cmidrule[\heavyrulewidth](lr){1-3}
\textbf{Dataset}     & \multicolumn{2}{c}{ImageNet}  \\ 
\cmidrule[\heavyrulewidth](lr){1-3}
Sparsity      & 90\%     & 80\%        \\ 
\cmidrule(lr){1-1}
\cmidrule(lr){2-3}
RigL+Sup-tickets(Ours)
&74.044 & 75.966
\\
RigL+Ensemble
&\textbf{74.074}	  &\textbf{76.022}	
\\
\cmidrule(lr){1-1}
\cmidrule(lr){2-3}
GraNet+Sup-tickets(Ours)
&74.554  & 76.168
\\
GraNet+Ensemble
&\textbf{74.614}	  & \textbf{76.198}
\\
\cmidrule[\heavyrulewidth](lr){1-3} 
\end{tabular} }
\end{table}

{Besides, we also apply knowledge distillation~\cite{hinton2015distilling} to distill the knowledge of three sup-tickets into a sparse student model. Each soft loss from the teacher model and the hard loss from the real label have equal weight in the final loss. Compared with knowledge distillation, we do not need to save all the sub-models as teacher models and do not need an extra round of training. Below we report the test accuracy of sparse VGG-16 on CIFAR-10/100. All the results are averaged from 3 random runs. Our method achieves higher accuracy (11 out of 12 cases) than the knowledge distillation based method.}

\begin{table*}[htbp]
\centering
\caption{{\textbf{Comparison with knowledge distillation.} Test accuracy (\%) of Sup-tickets and knowledge distillation (KD). In each setting, the best results are marked in bold.}}
\label{table:kd}
\resizebox{0.8\textwidth}{!}{
\begin{tabular}{lccc ccc}
\cmidrule[\heavyrulewidth](lr){1-7}

 \textbf{Dataset}     & \multicolumn{3}{c}{CIFAR-10} & \multicolumn{3}{c}{CIFAR-100}  \\ 
 \cmidrule(lr){1-1}
\cmidrule(lr){2-4}
\cmidrule(lr){5-7}

Sparsity     & 95\%      & 90\%     & 80\%     
     &  95\%      & 90\%     & 80\%         \\ 
     
 \cmidrule(lr){1-1}
\cmidrule(lr){2-4}
\cmidrule(lr){5-7}

SET+KD
& 93.13$\pm$0.06 & 93.56$\pm$0.16 & 93.53$\pm$0.10
& 70.73$\pm$0.18 & 71.79$\pm$0.42 & \textbf{73.06$\pm$0.02}
\\
SET+Sup-tickets (ours)
&\textbf{93.22$\pm$0.09}  & \textbf{93.63$\pm$0.05} & \textbf{93.80$\pm$0.13} 
&\textbf{71.18$\pm$0.29}  & \textbf{71.99$\pm$0.27} & {73.02$\pm$0.32}
\\
\cmidrule(lr){1-1}
\cmidrule(lr){2-4}
\cmidrule(lr){5-7}
RigL+KD
&92.98$\pm$0.15 & 93.38$\pm$0.14 & 93.61$\pm$0.15
&70.89$\pm$0.35 & 72.16$\pm$0.21 & 72.76$\pm$0.09
\\
RigL+Sup-tickets (ours)
&\textbf{93.20$\pm$0.13}  &  \textbf{93.81$\pm$0.11}  & \textbf{93.85$\pm$0.25}
&\textbf{71.31$\pm$0.21} & \textbf{72.57$\pm$0.29} &  {\textbf{73.61$\pm$0.11}}
\\
\cmidrule(lr){1-7}
\\

\end{tabular}}
\end{table*}

\newpage
\section{Statistical Significance}

{We analyze the statistical significance of the results obtained by Sup-tickets. To measure this, we perform Kolmogorov-Smirnov test~\citep{berger2014kolmogorov} (KS-test). The null hypothesis is that the two independent results/samples are drawn from the same continuous distribution. If the p-value is very small (p-value  \textless  0.05), it suggests that the difference between the two sets of results is significant, and the hypothesis is rejected. Otherwise, the obtained results are close together, and the hypothesis is true. We run the experiment on sparse VGG-16, CIFAR-10/100 for 15 runs with different random seeds and report the mean accuracy,  P-value, and decision of significance below. }

\begin{table*}[htbp]
\centering
\caption{{\textbf{Statistical Significance Analysis.}}}

\label{table:statistical_sig}
\resizebox{0.8\textwidth}{!}{
\begin{tabular}{lccc ccc}
\cmidrule[\heavyrulewidth](lr){1-7}

 \textbf{Dataset}     & \multicolumn{3}{c}{CIFAR-10} & \multicolumn{3}{c}{CIFAR-100}  \\ 
 \cmidrule(lr){1-1}
\cmidrule(lr){2-4}
\cmidrule(lr){5-7}

Sparsity     & 95\%      & 90\%     & 80\%     
     &  95\%      & 90\%     & 80\%         \\ 
     
 \cmidrule(lr){1-1}
\cmidrule(lr){2-4}
\cmidrule(lr){5-7}

SET
& 92.99 $\pm$0.16 & 93.41$\pm$0.20   & 93.65$\pm$0.15
& 70.50$\pm$0.31 & 71.55$\pm$0.38  & 72.76$\pm$0.21
\\
SET+Sup-tickets (ours)
&\textbf{93.17$\pm$0.16}  & \textbf{93.65$\pm$0.15} & \textbf{93.91$\pm$0.20} &
\textbf{71.18$\pm$0.27}  & \textbf{72.21$\pm$0.29} & \textbf{73.38$\pm$0.29}
\\
P-value 
& 5.90e-2 & 1.87e-2 & 1.02e-2
& 1.88e-05 & 1.02e-2 & 1.87e-2 

\\
Statistically significant  
& No &  Yes  & Yes
& Yes &  Yes &  Yes

\\
\cmidrule(lr){1-1}
\cmidrule(lr){2-4}
\cmidrule(lr){5-7}
RigL
& 92.94$\pm$0.20 & 93.41$\pm$0.14   &93.56$\pm$0.10
& 70.74$\pm$0.33   & 71.97$\pm$0.32  & 72.76$\pm$0.33 
\\
RigL+Sup-tickets (ours)
&\textbf{93.35$\pm$0.18}  &  \textbf{93.69$\pm$0.08}  & \textbf{93.85$\pm$0.12}
&\textbf{71.41$\pm$0.29} & \textbf{72.63$\pm$0.23} &  {\textbf{73.26$\pm$0.29}}

\\
P-value 
&1.63e-4 & 1.88e-05 & 1.88e-05
&1.02e-3 & 1.4e-06 & 1.87e-2

\\
Statistically significant  
& Yes &  Yes  & Yes
& Yes &  Yes &  Yes

\\
\cmidrule(lr){1-1}
\cmidrule(lr){2-4}
\cmidrule(lr){5-7}

\\

\end{tabular}}
\end{table*}

\section{comparison with SWA}

{Compared with SWA~\cite{izmailov2018averaging}, our approach provides two advantages. First of all, our method is much more training efficient as it only requires training a subset of the network during the whole training process. On the contrary, SWA requires to fully train a dense network even if it can be pruned afterward. Second, our method can efficiently discover and average \textit{multiple sparse sub-networks with different connectivity}, whereas SWA can only average sparse subnetworks with the same sparse connectivity. Different from dense neural networks where the connectivities are fixed, numerous sparse sub-networks with different connectivities are existing for sparse training, and all of them are capable of good performance. Instead of averaging sparse neural networks with the same sparse connectivity, it is more beneficial to average multiple sparse sub-networks with different connectivities since the sparse connectivity at initialization is insufficient to guarantee good performance.}

{Following we compare our method with two SWA-based methods. First, we run SWA with an additional step of pruning before the averaging. Unfortunately, it conflicts with the goal of sparse training, leading to more training FLOPs. In contrast, our approach follows a sparse-to-sparse paradigm that just trains a fraction of the parameters during the whole training process.  Second, we train a sparse model from scratch without considering connection exploration. The results below have empirically evaluated the benefits of our method that achieves better performance while requiring much fewer training FLOPs.}

\begin{table*}[htbp]
\centering
\caption{{\textbf{Comparison with SWA.} Test accuracy (\%) and training FLOPs of ResNet-50 on CIFAR100. The training FLOPs are normalized with the dense model. SWA baseline$^1$ means we train a dense model until the first averaging operation, prune it to the target sparsity with magnitude pruning, and then run SWA without exploring sparse connectivity. SWA baseline$^2$ indicates we initialize a model to certain sparse levels and perform SWA without connection exploration.  }}

\label{table:SWA}
\resizebox{0.9\textwidth}{!}{
\begin{tabular}{lccc ccc}
\cmidrule[\heavyrulewidth](lr){1-7}

 \textbf{Method}     & \multicolumn{3}{c}{Accuracy} & \multicolumn{3}{c}{Training FLOPs ( $\times 9.74e18$)}  \\ 
 \cmidrule(lr){1-1}
\cmidrule(lr){2-4}
\cmidrule(lr){5-7}

 & 95\% Sparsity      & 90\% Sparsity    & 80\% Sparsity     
 &  95\% Sparsity      & 90\% Sparsity    & 80\% Sparsity         \\ 
 
 \cmidrule(lr){1-1}
\cmidrule(lr){2-4}
\cmidrule(lr){5-7}
SWA baseline$^1$
& 76.64$\pm$0.45 &  77.23$\pm$0.44 & 77.72$\pm$0.29 
& $0.91\times$ & $0.92 \times$  & $0.93 \times$
\\
SWA baseline$^2$
&75.66$\pm$0.45 &  76.67$\pm$0.14 &  77.50$\pm$0.36 
& $0.11\times$ & $0.18 \times$  & $0.30 \times$
\\

Sup-tickets (ours)
&\textbf{77.58$\pm$0.47}  &  {\textbf{78.52$\pm$0.39}}  &    {\textbf{78.69$\pm$0.30}}
& $0.11\times$ & $0.18 \times$  & $0.30 \times$
\\
\cmidrule(lr){1-7}

\end{tabular}}
\end{table*}

\end{document}